# A Direct Estimation of High Dimensional Stationary Vector Autoregressions

Fang Han [*], Huanran Lu [†], and Han Liu [‡]

**Abstract**

The vector autoregressive (VAR) model is a powerful tool in learning complex time series and has been exploited in many fields. The VAR model poses some unique challenges to researchers: On one hand, the dimensionality, introduced by incorporating multiple numbers of time series and adding the order of the vector autoregression, is usually much higher than the time series length; On the other hand, the temporal dependence structure naturally present in the VAR model gives rise to extra difficulties in data analysis. The regular way in cracking the VAR model is via "least squares" and usually involves adding different penalty terms (e.g., ridge or lasso penalty) in handling high dimensionality. In this manuscript, we propose an alternative way in estimating the VAR model. The main idea is, via exploiting the temporal dependence structure, formulating the estimating problem to a linear program. There is instant advantage for the proposed approach over the lasso-type estimators: The estimation equation can be decomposed to multiple sub-equations and accordingly can be solved efficiently using parallel computing. Besides that, we also bring new theoretical insights into the VAR model analysis. So far the theoretical results developed in high dimensions (e.g., Song and Bickel (2011) and Kock and Callot (2012)) are based on stringent assumptions that are not transparent. Our results, on the other hand, show that the spectral norms of the transition matrices play an important role in estimation accuracy and build estimation and prediction consistency accordingly. Moreover, we provide some experiments on both synthetic and real-world equity data. We show that there are empirical advantages of our method over the lasso-type estimators in parameter estimation and forecasting.

## 1 Introduction

The vector autoregressive (VAR) model plays a fundamental role in analyzing multivariate time series data and has many applications in numerous academic fields. The VAR model is heavily used in finance (Tsay, 2005), econometrics (Sims, 1980), and brain imaging data analysis (Valdés-Sosa et al., 2005). For example, in understanding the brain connectivity network, multiple resting-state functional magnetic resonance imaging (rs-fMRI) data are obtained by consecutively scanning the same subject for approximately a hundred times or more. This naturally produces a high dimensional dependent data and a common strategy in handling such data is via building a vector autoregressive model (see Qiu et al. (2013) and the references therein).

[*]Department of Biostatistics, Johns Hopkins University, Baltimore, MD 21205, USA; e-mail: fhan@jhsph.edu

[†]Department of Operations Research and Financial Engineering, Princeton University, Princeton, NJ 08544, USA; e-mail: huanranl@princeton.edu.

[‡]Department of Operations Research and Financial Engineering, Princeton University, Princeton, NJ 08544, USA; e-mail: hanliu@princeton.edu.

This manuscript considers estimating the VAR model. Our focus is on the stationary vector autoregression with the order (or called lag) $p$ and Gaussian noises. More specifically, let random vectors $X_1, \ldots, X_T$ be from a stochastic process $(X_t)_{t=-\infty}^{\infty}$. Each $X_t$ is a $d$-dimensional random vector and satisfies that

$$X_t = \sum_{k=1}^{p} A_k^{\mathrm{T}} X_{t-k} + Z_t, \quad Z_t \sim N_d(0, \Psi), \tag{1.1}$$

where $A_1, \ldots, A_p$ are called the transition matrices and $(Z_t)_{t=-\infty}^{\infty}$ are independent multivariate Gaussian noises. Via assuming $\det(I_d - \sum_{k=1}^{p} A_k^{\mathrm{T}} z^k) \neq 0$ for all $z \in \mathcal{C}$ with modulus not greater than one, we then have the process is stationary (check, for example, Section 2.1 in Lütkepohl (2005)) and $X_t \sim N_d(0, \Sigma)$ for some covariance matrix $\Sigma$ depending on $\{A_k, k = 1, \ldots, p\}$ and $\Psi$.

There are in general three main targets in analyzing an VAR model. One is to estimate the transition matrices $A_1, \ldots, A_p$. These transition matrices reveal the temporal dependence in the data sequence and estimating them builds a fundamental first step in forecasting. Moreover, the zero and nonzero entries in the transition matrices directly incorporate the Granger non-causalities and causalities with regard to the stochastic sequence (see, for example, Corollary 2.2.1 in Lütkepohl (2005)). Another one of interest is the error covariance $\Psi$, which reveals the contemporaneous interactions among $d$ time series. Finally, by merely treating the temporal dependence as another measure of the data dependence (in parallel to the mixing conditions (Bradley, 2005)), it is also of interest to estimate the covariance matrix $\Sigma$.

This manuscript focuses on estimating the transition matrices $A_1, \ldots, A_p$, while noting that the techniques developed here can also be exploited to estimate the covariance matrix $\Sigma$ and the noise covariance $\Psi$. We first review the methods developed so far in transition matrix estimation. Let $A = (A_1^{\mathrm{T}}, \ldots, A_p^{\mathrm{T}})^{\mathrm{T}} \in \mathbb{R}^{dp \times d}$ be the combination of the transition matrices. Given $X_1, \ldots, X_T$, the perhaps most classic method in estimating $A$ is least squares minimization (Hamilton, 1994):

$$\widehat{A}^{\mathrm{LSE}} = \operatorname*{argmin}_{M \in \mathbb{R}^{dp \times d}} \|\widetilde{Y} - M^{\mathrm{T}} \widetilde{X}\|_{\mathsf{F}}^2, \tag{1.2}$$

where $\|\cdot\|_{\mathsf{F}}$ is the matrix Frobenius norm, $\widetilde{Y} = (X_{p+1}, \ldots, X_T) \in \mathbb{R}^{d \times (T-p)}$, and $\widetilde{X} = \{(X_p^{\mathrm{T}}, \ldots, X_1^{\mathrm{T}})^{\mathrm{T}}, \ldots, (X_{T-1}^{\mathrm{T}}, \ldots, X_{T-p}^{\mathrm{T}})^{\mathrm{T}}\} \in \mathbb{R}^{(dp) \times (T-p)}$. However, a fatal problem in (1.2) is that the product of the order of the autoregression $p$ and the number of time series $d$ is frequently larger than the time series length $T$. Therefore, the model has to be constrained to enforce identifiability. A common strategy is to add sparsity on the transition matrices so that the number of nonzero entries is less than $T$. Built on this assumption, there has been a large literature discussing adding different penalty terms to (1.2) for regularizing the estimator: From the ridge-penalty to the lasso-penalty and more non-concave penalty terms. In the following we list the major efforts. Hamilton (1994) discussed the use of the ridge penalty $\|M\|_{\mathsf{F}}^2$ in estimating the transition matrices. Hsu et al. (2008) proposed to add the $L_1$-penalty in estimating the transition matrices, inducing a sparse output. Several extensions to transition matrix estimation in the VAR model include: Wang et al. (2007) exploited the $L_1$-penalty in simultaneously estimating the regression coefficients and determining the number of lags in a linear regression model with autoregressive errors. In detecting causality, Haufe et al. (2008) transferred the problem to estimating transition matrices in an VAR model and advocated using a group lasso penalty for inducing joint sparsity among a whole block of coefficients. In studying the graphical Granger causality problem, Shojaie and Michailidis (2010) exploited the VAR model and proposed to estimate the coefficients using a truncated weighted $L_1$-penalty. Song and Bickel (2011) exploited the $L_1$ penalty in a complicated VAR model and aimed to select the variables and lags simultaneously.

The theoretical properties of the $L_1$-regularized estimator have been analyzed in Bento et al. (2010), Nardi and Rinaldo (2011), Song and Bickel (2011), and Kock and Callot (2012) under the assumption that the matrix $A$ is sparse, i.e., the number of nonzero entries in $A$ is much less than the dimension of parameters $pd^2$. Nardi and Rinaldo (2011) provided both subset and parameter estimation consistency results under a relatively low dimensional settings with $d = o(n^{1/2})$. Bento et al. (2010) studied the problem of estimating supports sets of the transition matrices in the high dimensional settings and proposed an "irrepresentable condition" similar as what is proposed in the linear regression model (Zou, 2006; Zhao and Yu, 2006; Meinshausen and Bühlmann, 2006; Wainwright, 2009). It is for the $L_1$ regularized estimator to attain the support set selection consistency. In parallel, Song and Bickel (2011) and Kock and Callot (2012) studied the parameter estimation and support set selection consistency of the $L_1$-regularized estimator in high dimensions.

In this paper, we propose a new approach to estimate the transition matrix $A$. Different from the line of lasso-based estimation procedures, which are built on penalizing the least square term, we exploit the linear programming technique and the proposed method is very fast to solve via parallel computing. Moreover, we do not need $A$ to be exactly sparse and allow it to be only "weakly sparse". The main idea is to estimate $A$ using the relationship between $A$ and the marginal and lag 1 autocovariance matrices (such a relationship is referred to as the Yule-Walker equation). We thus formulate the estimation procedure to a linear problem, while adding the $\|\cdot\|_{\max}$ (element-wise supremum norm) for model identifiability. Here we note that the proposed procedure can be considered as a generalization of the Dantzig selector (Candes and Tao, 2007) to the linear regression model with multivariate response. Indeed, our proposed method can also be exploited in conducting multivariate regression (Breiman and Friedman, 1997).

The proposed method enjoys several advantages compared to the existing ones: (i)

Computationally, our method can be formulated into $d$ linear programs and can be solved in parallel. Similar ideas have been used in learning high dimensional linear regression (Candes and Tao, 2007; Bickel et al., 2009) and graphical models (Yuan, 2010; Cai et al., 2011). (ii) In the model-level, our method allows $A$ to be only weakly sparse. (iii) Theoretically, so far the analysis on lasso-type estimators (Song and Bickel, 2011; Kock and Callot, 2012) depends on certain regularity conditions, restricted eigenvalue conditions on the design matrix for example, which are not transparent and do not explicitly reveal the role of temporal dependence in it. In contrast, we provide explicit nonasymptotic analysis, and our analysis highlights the spectral norm $\|A\|_2$ in estimation accuracy, which is inspired by some recent developments (Loh and Wainwright, 2012). Moreover, for exact sign recovery, our analysis does not need the "irrepresentable condition" which is usually required in the analysis of lasso type estimators (Bento et al., 2010).

The major theoretical results are briefly stated as follows. We adopt a double asymptotic framework where $d$ is allowed to increase with $T$. We call a matrix $s$-sparse if there are at most $s$ nonzero elements on each of its column. Under mild conditions, we provide the explicit rates of convergence of our estimator $\widehat{A}$ based on the assumption that $A$ is $s$-sparse (Cai et al., 2011). In particular, for lag 1 time series, we show that

$$\|\widehat{A} - A\|_1 = O_P\left\{\frac{s\|A\|_1}{1 - \|A\|_2}\left(\frac{\log d}{T}\right)^{1/2}\right\}, \quad \|\widehat{A} - A\|_{\max} = O_P\left\{\frac{\|A\|_1}{1 - \|A\|_2}\left(\frac{\log d}{T}\right)^{1/2}\right\},$$

where $\|\cdot\|_{\max}$ and $\|\cdot\|_q$ represent the matrix elementwise absolute maximum norm ($L_{\max}$ norm) and induced $L_q$ norm (detailed definitions will be provided in §2). Using the $L_{\max}$ norm consistency result, we further provide the sign recovery consistency of the proposed method. This result is of self interest and sheds light to detecting Granger causality. We also provide the prediction consistency results based on the $L_1$ consistency result and show that element-wise error in prediction can be controlled. Here for simplicity we only provide the results when $A$ is exactly sparse and defer the presentation of the results for weakly sparse matrix to Section 4.

The rest of the paper is organized as follows. In §2, we briefly review the vector autoregressive model. In §3, we introduce the proposed method for estimating the transition matrices of the vector autoregressive model. In §4, we provide the main theoretical results. In §5, we apply the new method to both synthetic and real equity data for illustrating its effectiveness. More discussions are provided in the last section. Detailed technical proofs are provided in the appendix[1].

## 2 Background

In this section, we briefly review the vector autoregressive model. Let $M = (M_{jk}) \in \mathbb{R}^{d \times d}$ and $v = (v_1, ..., v_d)^T \in \mathbb{R}^d$ be a matrix and an vector of interest. We denote $v_I$ to be the subvector of $v$ whose entries are indexed by a set $I \subset \{1, \ldots, d\}$. We also denote $M_{I,J}$ to be the submatrix of $M$ whose rows are indexed by $I$ and columns are indexed by $J$. We denote $M_{I,*}$ to be the submatrix of $M$ whose rows are indexed by $I$, $M_{*,J}$ to be the submatrix of $M$ whose columns are indexed by $J$. For $0 < q < \infty$, we define the $L_0$, $L_q$, and $L_\infty$ vector (pseudo-)norms to be

$$\|v\|_0 := \sum_{j=1}^{d} I(v_j \neq 0), \quad \|v\|_q := \Big(\sum_{j=1}^{d} |v_j|^q\Big)^{1/q}, \quad \text{and} \quad \|v\|_\infty := \max_{1 \leq j \leq d} |v_j|,$$

where $I(\cdot)$ is the indicator function. Letting $M$ be a matrix, we denote the matrix $L_q$, $L_{\max}$, and Frobenius ($L_\mathsf{F}$) norms to be

$$\|M\|_q := \max_{\|v\|_q=1} \|Mv\|_q, \quad \|M\|_{\max} := \max_{jk} |M_{jk}|, \quad \text{and} \quad \|M\|_\mathsf{F} := \Big(\sum_{j,k} |M_{jk}|^2\Big)^{1/2}.$$

---

[1] Some of the results in this paper were first stated without proof in a conference version (Han and Liu, 2013).

We denote $1_d = (1,\ldots,1)^\mathrm{T} \in \mathbb{R}^d$. Let $\sigma_1(M) \geq \cdots \geq \sigma_d(M)$ be the singular values of $M$.

Let $p \geq 1$ be an integer. A lag $p$ vector autoregressive process can be elaborated as follows: Let $(X_t)_{t=-\infty}^\infty$ be a stationary sequence of random vectors in $\mathbb{R}^d$ with mean 0 and covariance matrix $\Sigma$. We say that $(X_t)_{t=-\infty}^\infty$ follow a lag $p$ vector autoregressive model if and only if they satisfy

$$X_t = \sum_{k=1}^{p} A_k^\mathrm{T} X_{t-k} + Z_t \quad (t \in \mathbb{Z}). \tag{2.1}$$

Here $A_1,\ldots,A_p$ are called transition matrices. We denote $A = (A_1^\mathrm{T},\ldots,A_p^\mathrm{T})^\mathrm{T}$ to be the combination of the transition matrices. We assume that $Z_t$ are independently and identically generated from a Gaussian distribution $N_d(0,\Psi)$. Moreover, $Z_t$ and $(X_s)_{s<t}$ are independent for any $t \in \mathbb{Z}$. We pose an additional assumption that $\det(I_d - \sum_{k=1}^{p} A_k^\mathrm{T} z^k) \neq 0$ for all $z \in \mathcal{C}$ with modulus not greater than one. This guarantees that the sequence is stationary and we have, for any $t \in \mathbb{Z}$, $X_t$ follows a Gaussian distribution $N_d(0,\Sigma)$,

We denote $\Sigma_i(\cdot)$ to be an operator on the process $(X_t)_{t=-\infty}^\infty$. In particular, we define $\Sigma_i\{(X_t)\} = \mathrm{Cov}(X_0, X_i)$. It is easy to see that $\Sigma_0\{(X_t)\} = \Sigma$. If the lag of the vector autoregressive model is 1 (i.e., $X_t = A_1^\mathrm{T} X_{t-1} + Z_t$, for any $t \in \mathbb{Z}$), by simple calculation we have the so called "Yule-Walker Equation":

$$\Sigma_i\{(X_t)\} = \Sigma_0\{(X_t)\}(A_1)^i, \tag{2.2}$$

which further implies that

$$A_1 = [\Sigma_0\{(X_t)\}]^{-1} \cdot \Sigma_1\{(X_t)\}.$$

The results for lag 1 vector autoregressive model can be extended to the lag $p$ vector autoregressive model by appropriately redefining the random vectors. In detail, the autoregressive model with lag $p$ shown in (2.1) can be reformulated as an autoregressive model with lag 1:

$$\widetilde{X}_t = \widetilde{A}^\mathrm{T} \widetilde{X}_{t-1} + \widetilde{Z}_t, \tag{2.3}$$

where

$$\widetilde{X}_t = \begin{pmatrix} X_{t+p-1} \\ X_{t+p-2} \\ \vdots \\ X_t \end{pmatrix}, \quad \widetilde{A} = \begin{pmatrix} A_1 & I_d & 0 & \ldots & 0 \\ \vdots & \ddots & \ldots & \ldots & \vdots \\ A_{p-1} & 0 & 0 & \ldots & I_d \\ A_p & 0 & 0 & \ldots & 0 \end{pmatrix}, \quad \widetilde{Z}_t = \begin{pmatrix} Z_{t+p-1} \\ 0 \\ \vdots \\ 0 \end{pmatrix}. \quad (2.4)$$

Here $I_d \in \mathbb{R}^{d \times d}$ is the identity matrix, $\widetilde{X}_t \sim N_{dp}(0, \widetilde{\Sigma})$ for $t = 1, \ldots, T$, and $\widetilde{Z}_t \sim N_{dp}(0, \widetilde{\Psi})$ with $\widetilde{\Sigma} = \mathrm{Cov}(\widetilde{X}_t)$ and $\widetilde{\Psi} = \mathrm{Cov}(\widetilde{Z}_t)$. Therefore, we also have

$$\widetilde{A} = [\Sigma_0\{(\widetilde{X}_t)\}]^{-1} \cdot \Sigma_1\{(\widetilde{X}_t)\}. \quad (2.5)$$

This is similar to the relationship for the lag 1 vector autoregressive model.

## 3 Methods and Algorithms

We provide a new formulation to estimate $A_1, \ldots, A_p$ for the vector autoregressive model. Let $X_1, \ldots, X_T$ be from a lag $p$ vector autoregressive process $(X_t)_{t=-\infty}^{\infty}$ and we denote $\widetilde{X}_t = (X_{t+p-1}^{\mathrm{T}}, \ldots, X_t^{\mathrm{T}})^{\mathrm{T}}$ for $t = 1, \ldots, T - p + 1$. We denote $S$ and $S_1$ to be the marginal and lag 1 sample covariance matrices of $(\widetilde{X}_t)_{t=1}^{T-p+1}$:

$$S := \frac{1}{T-p+1} \sum_{t=1}^{T-p+1} \widetilde{X}_t \widetilde{X}_t^{\mathrm{T}}, \quad S_1 := \frac{1}{T-p} \sum_{t=1}^{T-p} \widetilde{X}_t \widetilde{X}_{t+1}^{\mathrm{T}}. \quad (3.1)$$

Using the connection between $\widetilde{A}$ and $\Sigma_0\{(\widetilde{X}_t)\}, \Sigma_1\{(\widetilde{X}_t)\}$ shown in (2.5), we know that a good estimator $\check{\Omega}$ of $\widetilde{A}$ shall satisfy that

$$\|\Sigma_0\{(\widetilde{X}_t)\}\check{\Omega} - \Sigma_1\{(\widetilde{X}_t)\}\| \quad (3.2)$$

is small enough with regard to a certain matrix norm $\|\cdot\|$. Moreover, using the fact that $A = (A_1^{\mathrm{T}}, \ldots, A_p^{\mathrm{T}})^{\mathrm{T}} = \widetilde{A}_{*,J}$, where $J = \{1, \ldots, d\}$, by (3.2) we have that a good estimate $\check{A}$ of $A$ shall satisfy

$$\|\Sigma_0\{(\widetilde{X}_t)\}\check{A} - [\Sigma_1\{(\widetilde{X}_t)\}]_{*,J}\| \quad (3.3)$$

is small enough.

Motivated by (3.3), we estimate $A_1, \ldots, A_p$ via replacing $\Sigma_0\{(\widetilde{X}_t)\}$ and $[\Sigma_1\{(\widetilde{X}_t)\}]_{*,J}$ with their empirical versions. For formulating the estimation equation to a linear program, we use the $L_{\max}$ norm. Accordingly, we end in solving the following convex optimization program:

$$\widehat{\Omega} = \operatorname*{argmin}_{M \in \mathbb{R}^{dp \times p}} \sum_{jk} |M_{jk}|, \quad \text{subject to} \quad \|SM - (S_1)_{*,J}\|_{\max} \leq \lambda_0, \quad (3.4)$$

where $\lambda_0 > 0$ is a tuning parameter. In (3.4), the constraint part aims to find an estimate that approximates the true parameter well, and combined with the minimization part, aims to induce certain sparsity. Let $\widehat{\Omega}_{*,j} = \widehat{\beta}_j$, it is easy to see that (3.4) can be decomposed to many subproblems and each $\widehat{\beta}_j$ can be solved by

$$\widehat{\beta}_j = \underset{v \in \mathbb{R}^d}{\operatorname{argmin}} \|v\|_1, \quad \text{subject to} \quad \|Sv - (S_1)_{*,j}\|_\infty \leq \lambda_0. \tag{3.5}$$

Accordingly, compared to the lasso-type procedures, the proposed method can be solved in parallel and therefore is computationally more efficient.

Once $\widehat{\Omega}$ is obtained, the estimator of the transition matrix $A_k$ can then be written as

$$\widehat{A}_k = \widehat{\Omega}_{J_k, *}, \tag{3.6}$$

where we denote $J_k = \{j : d(k-1) + 1 \leq j \leq dk\}$.

We now show that the optimization in (3.5) can be formulated into a linear program. Recall that any real number $a$ takes the decomposition $a = a^+ - a^-$, where $a^+ = a \cdot I(a \geq 0)$ and $a^- = -a \cdot I(a < 0)$. For any vector $v = (v_1, \ldots, v_d)^{\mathrm{T}} \in \mathbb{R}^d$, let $v^+ = (v_1^+, \ldots, v_d^+)^{\mathrm{T}}$ and $v^- = (v_1^-, \ldots, v_d^-)^{\mathrm{T}}$. We denote $v \geq 0$ if $v_1, \ldots, v_d \geq 0$ and $v < 0$ if $v_1, \ldots, v_d < 0$, $v_1 \geq v_2$ if $v_1 - v_2 \geq 0$, and $v_1 < v_2$ if $v_1 - v_2 < 0$. Letting $v = (v_1, \ldots, v_d)^{\mathrm{T}}$, the problem in (3.5) can be further relaxed to the following problem:

$$\widehat{\beta}_j = \underset{v^+, v^-}{\operatorname{argmin}} 1_d^{\mathrm{T}}(v^+ + v^-),$$

$$\text{subject to } \|Sv^+ - Sv^- - (S_1)_{*,j}\|_\infty \leq \lambda_0, \quad v^+ \geq 0, v^- \geq 0. \tag{3.7}$$

To minimize $1_d^{\mathrm{T}}(v^+ + v^-)$, $v^+$ or $v^-$ can not be both nonzero. Therefore, the solution to (3.7) is exactly the solution to (3.5). The optimization in (3.7) can be written as

$$\widehat{\beta}_j = \underset{v^+, v^-}{\operatorname{argmin}} 1_d^{\mathrm{T}}(v^+ + v^-),$$

$$\text{subject to} \quad Sv^+ - Sv^- - (S_1)_{*,j} \leq \lambda_0 1_d,$$

$$-Sv^+ + Sv^- + (S_1)_{*,j} \leq \lambda_0 1_d,$$

$$v^+ \geq 0, v^- \geq 0.$$

This is equivalent to

$$\widehat{\beta}_j = \underset{\omega}{\operatorname{argmin}} 1_{2d}^{\mathrm{T}} \omega, \quad \text{subject to} \quad \theta + W\omega \geq 0, \quad \omega \geq 0, \tag{3.8}$$

where

$$\omega = \begin{pmatrix} v^+ \\ v^- \end{pmatrix}, \quad \theta = \begin{bmatrix} (S_1)_{*,j} + \lambda_0 1_d \\ -(S_1)_{*,j} + \lambda_0 1_d \end{bmatrix}, \quad W = \begin{pmatrix} -S & S \\ S & -S \end{pmatrix}.$$

The optimization (3.8) is a linear program. We can solve it using the simplex algorithm (Murty, 1983).

## 4 Theoretical Properties

In this section, under the double asymptotic framework, we provide the nonasymptotic rates of convergence in parameter estimation under the matrix $L_1$ and $L_{\max}$ norms.

We first present the rates of convergence of the estimator $\widehat{\Omega}$ in (3.4) under the vector autoregressive model with lag 1. This result allows us to sharply characterize the impact of the temporal dependence of the time series on the obtained rate of convergence. In particular, we show that the rate of convergence is closely related to the $L_1$ and $L_2$ norms of the transition matrix $A_1$, where $\|A_1\|_2$ is the key part in characterizing the impact of temporal dependence on estimation accuracy. Secondly, we present the sign recovery consistency result of our estimator. Compared to the lasso-type estimators, our result does not require the irrepresentable condition. These results are combined together to show that we have the prediction consistency, i.e., the term $\|A_1 X_T - \widehat{A}_1 X_T\|$ goes to zero with regard to certain norms $\|\cdot\|$. In the end, we extend these results from the vector autoregressive model with lag 1 to lag $p$ with $p > 1$.

We start with some additional notation. Let $M_d \in \mathbb{R}$ be a quantity which may scale with the time series length and dimension $(T, d)$. We define the set of square matrices in $\mathbb{R}^{d \times d}$, denoted by $\mathcal{M}(q, s, M_d)$, as

$$\mathcal{M}(q, s, M_d) := \Big\{ M \in \mathbb{R}^{d \times d} : \max_{1 \leq j \leq d} \sum_{i=1}^{d} |M_{ij}|^q \leq s, \|M\|_1 \leq M_d \Big\}.$$

For $q = 0$, the class $\mathcal{M}(0, s, M_d)$ contains all the $s$-sparse matrices with bounded $L_1$ norms.

There are two general remarks about the model $\mathcal{M}(q, s, M_d)$: (i) $\mathcal{M}(q, s, M_d)$ can be considered as the matrix version of the vector "weakly sparse set" explored in Raskutti et al. (2011) and Vu and Lei (2012). Such a way to define the weakly sparse set of matrices is also investigated in Cai et al. (2011). (ii) For the exactly sparse matrix set, $\mathcal{M}(0, s, M_d)$, the sparsity level $s$ here represents the largest number of nonzero entries in each column of the matrix. In contrast, the sparsity level $s'$ exploited in Kock and Callot (2012) is the total number of nonzero entries in the matrix. We must have $s' \geq s$ and regularly $s' \gg s$ (means $s/s' \to 0$).

The next theorem presents the $L_1$ and $L_{\max}$ rates of convergence of our estimator under the vector autoregressive model with lag 1.

**Theorem 4.1.** *Suppose that $(X_t)_{t=1}^T$ are from a lag 1 vector autoregressive process $(X_t)_{t=-\infty}^{\infty}$ as described in (2.1). We assume the transition matrix $A_1 \in \mathcal{M}(q, s, M_d)$ for some $0 \leq q < 1$. Let $\widehat{A}_1$ be the optimum to (3.4) with the tuning parameter*

$$\lambda_0 = \frac{32\|\Sigma\|_2 \max_j(\Sigma_{jj})}{\min_j(\Sigma_{jj})(1 - \|A\|_2)} (2M_d + 3) \left( \frac{\log d}{T} \right)^{1/2}.$$

*For $T \geq 6 \log d + 1$ and $d \geq 8$, we have, with probability no smaller than $1 - 14d^{-1}$,*

$$\|\widehat{A}_1 - A_1\|_1 \leq 4s \left\{ \frac{32\|\Sigma^{-1}\|_1 \max_j(\Sigma_{jj}) \|\Sigma\|_2}{\min_j(\Sigma_{jj})(1 - \|A_1\|_2)} (2M_d + 3) \left( \frac{\log d}{T} \right)^{1/2} \right\}^{1-q}. \tag{4.1}$$

Moreover, with probability no smaller than $1 - 14d^{-1}$,

$$\|\widehat{A}_1 - A_1\|_{\max} \leq \frac{64\|\Sigma^{-1}\|_1 \max_j(\Sigma_{jj})\|\Sigma\|_2}{\min_j(\Sigma_{jj})(1 - \|A_1\|_2)}(2M_d + 3)\left(\frac{\log d}{T}\right)^{1/2}. \quad (4.2)$$

In the above results, $\Sigma$ is the marginal covariance matrix of $X_t$.

It can be observed that, similar to the lasso and Dantzig selector (Candes and Tao, 2007; Bickel et al., 2009), the tuning parameter $\lambda_0$ here depends on the variance term $\Sigma$. In practice, same as most preceded developments (see, for example, Song and Bickel (2011)), we can use a data-driven way to select the tuning parameter. In this manuscript we explore using cross-validation to choose $\lambda_0$ with the best prediction accuracy. In Section 5 we will show that the procedure of selecting the tuning parameter via cross-validation gives reasonable results.

Here $A_1$ is assumed to be at least weakly sparse and belong to the set $\mathcal{M}(q, s, M_d)$. This is merely for the purpose of model identifiability. Otherwise, we will have multiple global optima in the optimization problem.

The obtained rates of convergence in Theorem 4.1 depend on both $\Sigma$ and $A_1$ with $\|A_1\|_2$ characterizing the temporal dependence. In particular, the estimation error is related to the spectral norm of the transition matrix $A_1$. Intuitively, this is because $\|A_1\|_2$ characterizes the data dependence of $X_1, \ldots, X_T$, and accordingly intrinsically characterizes how much information there is in the data. If $\|A_1\|_2$ is larger, then there is less information we can exploit in estimating $A_1$. Technically, $\|A_1\|_2$ determines the rate of convergence of $S$ and $S_1$ to their population counterparts. We refer to the proofs of Lemmas A.1 and A.2 for details.

In the following, we list two examples to provide more insights about the results in Theorem 4.1.

**Example 4.2.** *We consider the case where $\Sigma$ is a strictly diagonal dominant (SDD) matrix (Horn and Johnson, 1990) with the property*

$$\delta_i := |\Sigma_{ii}| - \sum_{j \neq i}|\Sigma_{ij}| \geq 0, \quad (i = 1, \ldots, d).$$

*This corresponds to the cases where the $d$ entries in any $X_t$ with $t \in \{1, \ldots, T\}$ are weakly dependent. In this setting, Ahlberg and Nilson (1963) showed that*

$$\|\Sigma^{-1}\|_1 = \|\Sigma^{-1}\|_\infty \leq \left\{\min_i\left(|\Sigma_{ii}| - \sum_{j \neq i}|\Sigma_{ij}|\right)\right\}^{-1} = \max_i(\delta_i^{-1}). \quad (4.3)$$

*Moreover, by algebra, we have*

$$\|\Sigma\|_2 \leq \|\Sigma\|_1 = \max_i\left(|\Sigma_{ii}| + \sum_{j \neq i}|\Sigma_{ij}|\right) \leq 2\max_i(|\Sigma_{ii}|). \quad (4.4)$$

Equations (4.3) and (4.4) suggest that, when $\max_i(\Sigma_{ii})$ is upper bounded, and both $\min_i(\Sigma_{ii})$ and $\delta_i$ are lower bounded by a fixed constant, we have both $\|\Sigma^{-1}\|_1$ and $\|\Sigma\|_2$ are upper bounded, and the obtained rates of convergence in (4.1) and (4.2) can be simplified as:

$$\|\widehat{A}_1 - A_1\|_1 = O_P\left[s\left\{\frac{M_d}{1-\|A_1\|_2}\left(\frac{\log d}{T}\right)^{1/2}\right\}^{1-q}\right],$$

$$\|\widehat{A}_1 - A_1\|_{\max} = O_P\left\{\frac{M_d}{1-\|A_1\|_2}\left(\frac{\log d}{T}\right)^{1/2}\right\}.$$

**Example 4.3.** *We can generalize the "entry-wise weakly dependent" structure in Example 4.2 to a "block-wise weakly dependent" structure. More specifically, we consider the case where $\Sigma = (\Sigma^b_{jk})$ with blocks $\Sigma^b_{jk} \in \mathbb{R}^{d_j \times d_k}$ ($1 \le j \le K$) is a strictly block diagonal dominant (SBDD) matrix with the property*

$$\delta^b_i = \|(\Sigma^b_{ii})^{-1}\|_\infty^{-1} - \sum_{j \ne i} \|\Sigma^b_{ij}\|_\infty > 0 \quad (i = 1, \ldots, K).$$

*In this case, Varah (1975) showed that*

$$\|\Sigma^{-1}\|_1 = \|\Sigma^{-1}\|_\infty \le \left\{\min_i\left(\|(\Sigma^b_{ii})^{-1}\|_\infty^{-1} - \sum_{j \ne i}\|\Sigma^b_{ij}\|_\infty\right)\right\}^{-1} = \max\{(\delta^b_i)^{-1}\}.$$

*Moreover, we have*

$$\|\Sigma\|_2 \le \|\Sigma\|_1 \le \max_i(\|(\Sigma^b_{ii})^{-1}\|_\infty^{-1} + \|\Sigma^b_{ii}\|_\infty).$$

*Accordingly, generally $(\|(\Sigma^b_{ii})^{-1}\|_\infty^{-1} + \|\Sigma^b_{ii}\|_\infty)$ is in the scale of $\max_i(d_i) \ll d$, and when $\delta^b_i$ are lower bounded and the condition number of $\Sigma$ is upper bounded, we have the obtained rates of convergence can be simplified as:*

$$\|\widehat{A}_1 - A_1\|_1 = O_P\left[s\left\{\frac{M_d \cdot \max_i(d_i)}{1-\|A_1\|_2}\left(\frac{\log d}{T}\right)^{1/2}\right\}^{1-q}\right],$$

$$\|\widehat{A}_1 - A_1\|_{\max} = O_P\left\{\frac{M_d \cdot \max_i(d_i)}{1-\|A_1\|_2}\left(\frac{\log d}{T}\right)^{1/2}\right\}.$$

We then continue to the results of feature selection. If we have $A_1 \in \mathcal{M}(0, s, M_d)$, from the element-wise $L_{\max}$ norm convergence, a sign recovery result can be obtained. In detail, let $\check{A}_1$ be a truncated version of $\widehat{A}_1$ with level $\gamma$:

$$(\check{A}_1)_{ij} = (\widehat{A}_1)_{ij} I\{|(\widehat{A}_1)_{ij}| \ge \gamma\}. \tag{4.5}$$

The following corollary shows that $\check{A}_1$ recovers the sign of $A_1$ with overwhelming probability.

**Corollary 4.4.** *Suppose that the conditions in Theorem 4.1 hold and $A_1 \in \mathcal{M}(0, s, M_d)$. If we choose the truncation level*

$$\gamma = \frac{64\|\Sigma^{-1}\|_1 \max_j(\Sigma_{jj})\|\Sigma\|_2}{\min_j(\Sigma_{jj})(1 - \|A_1\|_2)}(2M_d + 3)\left(\frac{\log d}{T}\right)^{1/2}$$

*in (4.5) and with the assumption that*

$$\min_{\{(j,k):(A_1)_{jk} \neq 0\}} |(A_1)_{jk}| \geq 2\gamma,$$

*we have, with probability no smaller than $1 - 14d^{-1}$, $\operatorname{sign}(A_1) = \operatorname{sign}(\check{A}_1)$. Here for any matrix $M$, $\operatorname{sign}(M)$ is a matrix with each element representing the sign of the corresponding entry in $M$.*

Here we note that Corollary 4.4 sheds lights to detecting Granger causality. For any two processes $\{y_t\}$ and $\{z_t\}$, Granger (1969) defined the causal relationship in principle as follows: Provided that we know everything in the universe, $\{y_t\}$ is said to cause $\{z_t\}$ in Granger's sense if removing the information about $\{y_s\}_{s \leq t}$ from the whole knowledge base built by time $t$ will increase the prediction error about $z_t$. It is known that the noncausalities are determined by the transition matrices in the stable VAR process (Lütkepohl, 2005). Therefore, detecting the nonzero entries of $A_1$ consistently means that we can estimate the Granger-causality network consistently.

We then turn to evaluate the prediction performance of the proposed method. Given a new data point $X_{T+1}$ in the time point $T+1$, based on $(X_t)_{t=1}^T$, the next corollary quantifies the distance between $X_{T+1}$ and $\widehat{A}_1 X_T$ in terms of $L_\infty$ norm.

**Corollary 4.5.** *Suppose that the conditions in Theorem 4.1 hold and let*

$$\Psi_{\max} := \max_i(\Psi_{ii}) \quad \text{and} \quad \Sigma_{\max} := \max_i(\Sigma_{ii}).$$

*Then for the new data point $X_{T+1}$ at time point $T+1$ and any constant $\alpha > 0$, with probability greater than*

$$1 - 2(d^{\alpha/2-1}\sqrt{\pi/2 \cdot \alpha \log d\})^{-1} - 14d^{-1},$$

*we have*

$$\|X_{T+1} - \widehat{A}_1^\mathrm{T} X_T\|_\infty \leq (\Psi_{\max} \cdot \alpha \log d)^{1/2} +$$

$$4s\left\{\frac{32\|\Sigma^{-1}\|_1 \max_j(\Sigma_{jj})\|\Sigma\|_2}{\min_j(\Sigma_{jj})(1 - \|A_1\|_2)}(2M_d + 3)\left(\frac{\log d}{T}\right)^{1/2}\right\}^{1-q} \cdot (\Sigma_{\max} \cdot \alpha \log d)^{1/2}, \quad (4.6)$$

*where $\widehat{A}_1$ is calculated based on $(X_t)_{t=1}^T$.*

Here we note that the first term in the right-hand side of Equation (4.6), $(\Psi_{\max} \cdot \alpha \log d)^{1/2}$, is present due to the diverges of the new data point from its mean caused by an unpredictable noise perturb term $Z_{T+1} \sim N_d(0, \Psi)$. This term is unable to be canceled out even if we have almost infinite data points. The second term in the right-hand side of Equation (4.6) depends on the estimation accuracy of $\widehat{A}_1$ to $A_1$ and will converge to zero under certain conditions. In other words, the term

$$\|A_1^\mathrm{T} X_T - \widehat{A}_1^\mathrm{T} X_T\|_\infty \to 0, \tag{4.7}$$

converges to zero in probability as $n, d \to \infty$.

Although $A_1$ is in general asymmetric, there exist cases such that a symmetric transition matrix is more of interest. It is known that the off-diagonal entries in the transition matrix represent the influence of one state on the others and such influence might be symmetric or not. Weiner et al. (2012) provided several examples where a symmetric transition matrix is more appropriate for modeling the data.

If we can further suppose that the transition matrix $A_1$ is symmetric, we can use this information and obtain a new estimator $\bar{A}_1$ as

$$(\bar{A}_1)_{jk} = (\bar{A}_1)_{kj} := (\widehat{A}_1)_{jk} I(|(\widehat{A}_1)_{jk}| \leq |(\widehat{A}_1)_{kj}|) + (\widehat{A}_1)_{kj} I(|(\widehat{A}_1)_{kj}| \leq |(\widehat{A}_1)_{jk}|).$$

In other word, we always pick the entry with smaller magnitudes. Then using Theorem 4.1, we have $\|\bar{A}_1 - A_1\|_1$ and $\|\bar{A}_1 - A_1\|_\infty$ can be upper bounded by the same number presented in the right-hand side of (4.1). In this case, because both $A_1$ and $\bar{A}_1$ are symmetric, we have $\|\bar{A}_1 - A_1\|_2 \leq \|\bar{A}_1 - A_1\|_1 = \|\bar{A}_1 - A_1\|_\infty$. We then proceed to quantify the prediction accuracy under $L_2$ norm in the next corollary.

**Corollary 4.6.** *Suppose that the conditions in Theorem 4.1 hold and $A_1$ is a symmetric matrix. Then for the new data point $X_{T+1}$ at time point $T+1$, with probability greater than $1 - 18d^{-1}$, we have*

$$\|X_{T+1} - \bar{A}_1^\mathrm{T} X_T\|_2 \leq \sqrt{2\|\Psi\|_2 \log d} + \sqrt{\mathrm{tr}(\Psi)} +$$

$$4s \left\{ \frac{32\|\Sigma^{-1}\|_1 \max_j(\Sigma_{jj}) \|\Sigma\|_2}{\min_j(\Sigma_{jj})(1 - \|A_1\|_2)} (2M_d + 3) \left(\frac{\log d}{T}\right)^{1/2} \right\}^{1-q} \cdot \{\sqrt{2\|\Sigma\|_2 \log d} + \sqrt{\mathrm{tr}(\Sigma)}\}. \tag{4.8}$$

Based on Corollary 4.6, we have, similar as what is discussed in Corollary 4.5, the term $\|A_1^\mathrm{T} X_T - \widehat{A}_1^\mathrm{T} X_T\|_2$ will vanish when the second term in the left-hand side of (4.8) can converge to zero.

Using the augmented formulation of the lag $p$ vector autoregressive model in (2.3), we can extend the results in Theorem 4.1 from lag 1 to the more general lag $p$ model with $p \geq 1$.

**Theorem 4.7.** *Suppose that $(X_t)_{t=1}^T$ are from a lag $p$ vector autoregressive process $(X_t)_{t=-\infty}^\infty$ as described in (2.1). Let $\widetilde{A}$ and $\widetilde{\Sigma}$ be defined as in §2. We assume that $\widetilde{A} \in \mathcal{M}(q, s, M_{dp})$ for some $0 \leq q < 1$. Let $\widehat{\Omega}$ be the optimum to (3.4) with tuning parameter*

$$\lambda_0 = \frac{C\|\widetilde{\Sigma}\|_2 \max_j(\widetilde{\Sigma}_{jj}) \max(M_{dp}, 1)}{\min_j(\widetilde{\Sigma}_{jj})(1 - \|\widetilde{A}\|_2)} \left(\frac{\log d + \log p}{T - p}\right)^{1/2},$$

*where $C$ is a generic constant. Then we have,*

$$\sum_{k=1}^{p} \|\widehat{A}_k - A_k\|_1 = \|\widehat{\Omega} - A\|_1 = O_P\left[s\left\{\frac{\|\widetilde{\Sigma}\|_2 \max_j(\widetilde{\Sigma}_{jj}) \max(M_{dp}, 1)}{\min_j(\widetilde{\Sigma}_{jj})(1 - \|\widetilde{A}\|_2)} \left(\frac{\log d + \log p}{T - p}\right)^{1/2}\right\}^{1-q}\right],$$

$$\max_k \|\widehat{A}_k - A_k\|_{\max} = \|\widehat{\Omega} - A\|_{\max} = O_P\left\{\frac{\|\widetilde{\Sigma}\|_2 \max_j(\widetilde{\Sigma}_{jj}) \max(M_{dp}, 1)}{\min_j(\widetilde{\Sigma}_{jj})(1 - \|\widetilde{A}\|_2)} \left(\frac{\log d + \log p}{T - p}\right)^{1/2}\right\}.$$

*Here we remind that $A = (A_1^{\mathrm{T}}, \ldots, A_p^{\mathrm{T}})^{\mathrm{T}}$ and $\widehat{A}_k$ is defined in (3.6) for $k = 1, \ldots, p$.*

With the augmented formulation, we have that similar arguments as shown in Corollaries 4.4, 4.5, and 4.6 also hold.

## 5 Experiments

We conduct numerical experiments on both synthetic and real data to illustrate the effectiveness of our proposed method compared to the competing ones, as well as obtain more insights on the performance of the proposed method. In the following we consider the three competing methods:

- (i) The least square estimation using a ridge penalty (The method in Hamilton (1994) by adding a ridge penalty $\|M\|_F^2$ to the least squares loss function in (1.2)).

- (ii) The least square estimation using an $L_1$ penalty (The method in Hsu et al. (2008) by adding an $L_1$ penalty $\sum_{ij} |M_{ij}|$ to (1.2)).

- (iii) Our method (The estimator described in (3.4)).

Here we consider including the procedure discussed in Hamilton (1994) because it is a commonly explored baseline and shows how bad the classic procedure can be when the dimension is high. We only consider the competing procedure proposed in Hsu et al. (2008) because this is the only method that is specifically designed for the same simple VAR as what we study. We do not consider other aforementioned procedures (e.g.,Haufe et al. (2008), Shojaie and Michailidis (2010)) because they are designed for more specific models with more assumptions. We use the R package "glmnet" (Friedman et al., 2009) for implementing the lasso method in Hsu et al. (2008), and the simplex algorithm for implementing ours.

### 5.1 Cross-Validation Procedure

We start with an introduction to how to conduct cross-validation for choosing the lag $p$ and the tuning parameter $\lambda$ in the algorithm outlined in Section 3.

For the time series $(X_t)_{t=-\infty}^{T}$ and a specific time point $t_0$ of interest, if both $p$ and $\lambda$ are assumed to be unknown, the proposed cross-validation procedure is as follows.

1. We set all possible choices of $(p, \lambda)$ to be a grid. We set $n_1$ and $n_2$ to be two numbers (representing the length of training data and the number of replicates).

2. For each $X_t$ among $X_{t_0-1}, \ldots, X_{t_0-n_2}$, the estimates $\widehat{A}_1^t(p, \lambda), \ldots, \widehat{A}_p^t(p, \lambda)$ are calculated based on the training data $X_{t-1}, \ldots, X_{t-n_1}$ and any choice of $(p, \lambda)$. We set the prediction error at time $t$, denoted as $\mathrm{Err}_t(p, \lambda)$, to be $\mathrm{Err}_t(p, \lambda) := \|X_t - \sum_{k=1}^p \widehat{A}_k^t(p, \lambda)^\mathrm{T} X_{t-k}\|_2$.

3. We take an average over the prediction errors and denote

$$\overline{\mathrm{Err}}(p, \lambda) := \frac{1}{n_2} \sum_{t=t_0-n_2}^{t_0-1} \mathrm{Err}_t(p, \lambda)$$

.

4. We choose the $(p, \lambda)$ over the grid such that $\overline{\mathrm{Err}}(p, \lambda)$ is minimized.

In case when $p$ is predetermined, the above procedure can be easily modified to focus only on selecting $\lambda$ with $p$ to be the determined value.

## 5.2 Synthetic Data Analysis

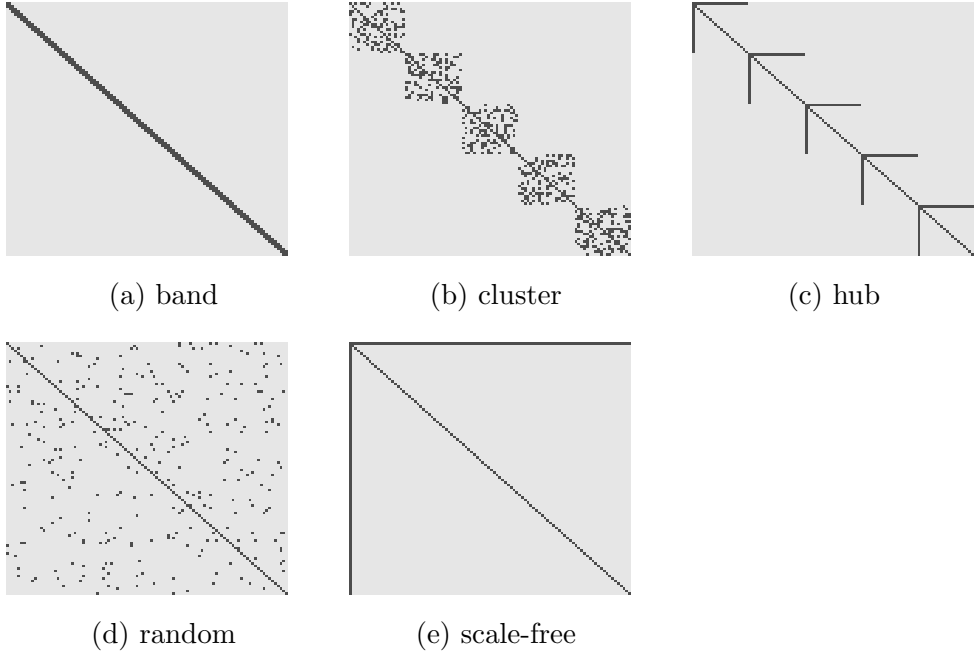

(a) band  (b) cluster  (c) hub

(d) random  (e) scale-free

Figure 1: Five different transition matrix patterns used in the experiments. Here gray points represent the zero entries and black points represent nonzero entries.

### 5.2.1 Performance Comparison: Lag $p = 1$

This section focuses on vector autoregressive model described in (2.1) with lag one. We compare our method to the competing ones on several synthetic datasets. We consider the settings where the time series length $T$ varies from 50 to 100 and the dimension $d$ varies from 50 to 200.

We create the transition matrix $A_1$ according to five different patterns: band, cluster, hub, random, and scale-free. Typical realizations of these patterns are illustrated in Figure 1 and are generated using the "flare" package in R (Li et al., 2013). In those plots, the gray points represent the zero entries and the black points represent the nonzero entries. We then rescale $A_1$ such that we have $\|A_1\|_2 = 0.5$. Once $A_1$ is obtained, we generate $\Sigma$ using two models. First is the simple setting with $\Sigma$ to be diagonal:

$$\Sigma = 2\|A_1\|_2 I_d. \tag{5.1}$$

The second is the complex setting where $\Sigma$ is of Toeplitz form:

$$\Sigma_{i,i} = 1, \quad \Sigma_{i,j} = \rho^{|i-j|} \text{ for some } \rho \in (0,1) \text{ and } i,j = 1,\ldots,d.$$

We then calculate the covariance matrix $\Psi$ of the Gaussian noise vector $Z_t$ as $\Psi = \Sigma - A_1^{\mathrm{T}} \Sigma A_1$. With $A_1, \Sigma$, and $\Psi$, we simulate a time series $(X_1, \ldots, X_T)^{\mathrm{T}} \in \mathbb{R}^{T \times d}$ according to the model described in (2.1).

We construct 1,000 replicates and compare the three methods described above. The averaged estimation errors under different matrix norms are illustrated in Tables 1 to 10. The standard deviations of the estimation errors are provided in the parentheses. The tuning parameters for the three methods are selected using the cross-validation procedure outlined in Section 5.1 with $n_1 = T/2$, $n_2 = T/2$, and the lag $p$ predetermined to be 1.

Tables 1 to 10 show that our method nearly uniformly outperforms the methods in Hsu et al. (2008) and Hamilton (1994) under different norms (Frobenius, $L_2$, and $L_1$ norms). In particular, the improvement over the method in Hsu et al. (2008) tends to be more significant when the dimension $d$ is larger. Our method also has averagely slightly less standard deviations compared to the method in Hsu et al. (2008), but overall the difference is not significant. The method in Hamilton (1994) has worse performance than the other two methods. This verifies that it is not appropriate to handle very high dimensional data.

### 5.2.2 Synthetic Data: Lag $p \geq 1$

In this section, we further compare the performance of the three competing methods under the settings of possibly multiple lags, with the number of lags known.

In detail, we choose $p$ to be from 1 to 9, the time series length $T = 100$, and the dimension $d = 50$. The transition matrices $A_1, \ldots, A_p$ are created according to "hub" or "scale-free" pattern, and then rescaled such that $\|A_i\|_2 = 0.1$ for $i = 1, \ldots, p$. The error covariance matrix $\Psi$ is set to be identity for simplicity. Under this multiple lags setting, we then calculate the covariance matrix of $\widetilde{X}_t$, i.e. $\widetilde{\Sigma}$ defined in (2.4), by solving a discrete Lyapunov equation $\widetilde{A}^{\mathrm{T}} \widetilde{\Sigma} \widetilde{A} - \widetilde{\Sigma} + \widetilde{\Psi} = 0$. This is via using the Matlab command "dlyapchol".

Table 1: Comparison of estimation performance of three methods with diagonal covariance matrix over 1,000 replications. The standard deviations are presented in the parentheses. Here $L_{\mathsf{F}}, L_2$, and $L_1$ represent the Frobenius, $L_2$, and $L_1$ matrix norms respectively. The pattern of the transition matrix is "band".

| | | ridge method | | | lasso method | | | our method | | |
|---|---|---|---|---|---|---|---|---|---|---|
| $d$ | $T$ | $L_{\mathsf{F}}$ | $L_2$ | $L_1$ | $L_{\mathsf{F}}$ | $L_2$ | $L_1$ | $L_{\mathsf{F}}$ | $L_2$ | $L_1$ |
| 50 | 100 | 2.71 | 0.52 | 2.47 | 2.34 | 0.50 | 1.54 | **2.08** | **0.49** | **0.58** |
| | | (0.028) | (0.023) | (0.103) | (0.064) | (0.029) | (0.161) | (0.045) | (0.006) | (0.039) |
| 100 | 50 | 4.21 | 0.64 | 3.54 | 5.52 | 0.75 | 3.13 | **3.26** | **0.52** | **1.03** |
| | | (0.026) | (0.024) | (0.136) | (0.075) | (0.024) | (0.211) | (0.052) | (0.017) | (0.321) |
| 200 | 100 | 7.28 | 0.76 | 6.26 | 6.36 | 0.64 | 2.77 | **4.26** | **0.50** | **0.69** |
| | | (0.031) | (0.018) | (0.132) | (0.057) | (0.015) | (0.112) | (0.045) | (0.003) | (0.035) |

Table 2: Comparison of estimation performance of three methods with diagonal covariance matrix over 1,000 replications. The standard deviations are presented in the parentheses. Here $L_{\mathsf{F}}, L_2$, and $L_1$ represent the Frobenius, $L_2$, and $L_1$ matrix norms respectively. The pattern of the transition matrix is "cluster".

| | | ridge method | | | lasso method | | | our method | | |
|---|---|---|---|---|---|---|---|---|---|---|
| $d$ | $T$ | $L_{\mathsf{F}}$ | $L_2$ | $L_1$ | $L_{\mathsf{F}}$ | $L_2$ | $L_1$ | $L_{\mathsf{F}}$ | $L_2$ | $L_1$ |
| 50 | 100 | 2.48 | 0.44 | 2.40 | 2.12 | **0.43** | 1.56 | **1.48** | 0.49 | **0.69** |
| | | (0.034) | (0.024) | (0.110) | (0.055) | (0.032) | (0.119) | (0.020) | (0.011) | (0.026) |
| 100 | 50 | 3.74 | 0.58 | 3.46 | 5.24 | 0.67 | 3.16 | **2.27** | **0.50** | **0.66** |
| | | (0.031) | (0.022) | (0.121) | (0.084) | (0.025) | (0.223) | (0.002) | (0.001) | (0.002) |
| 200 | 100 | 6.80 | 0.72 | 6.26 | 5.82 | 0.55 | 2.80 | **3.02** | **0.49** | **0.77** |
| | | (0.025) | (0.021) | (0.188) | (0.058) | (0.014) | (0.109) | (0.024) | (0.010) | (0.047) |

Table 3: Comparison of estimation performance of three methods with diagonal covariance matrix over 1,000 replications. The standard deviations are presented in the parentheses. Here $L_\mathsf{F}, L_2$, and $L_1$ represent the Frobenius, $L_2$, and $L_1$ matrix norms respectively. The pattern of the transition matrix is "hub".

|     |     | ridge method | | | lasso method | | | our method | | |
| --- | --- | --- | --- | --- | --- | --- | --- | --- | --- | --- |
| $d$ | $T$ | $L_\mathsf{F}$ | $L_2$ | $L_1$ | $L_\mathsf{F}$ | $L_2$ | $L_1$ | $L_\mathsf{F}$ | $L_2$ | $L_1$ |
| 50 | 100 | 2.41 | 0.42 | 2.37 | 1.96 | **0.38** | 1.48 | **1.16** | 0.41 | **1.05** |
|    |     | (0.033) | (0.027) | (0.102) | (0.06) | (0.039) | (0.141) | (0.115) | (0.058) | (0.092) |
| 100 | 50 | 3.49 | 0.55 | 3.44 | 5.06 | 0.63 | 3.11 | **1.86** | **0.50** | **1.40** |
|    |     | (0.034) | (0.023) | (0.143) | (0.088) | (0.032) | (0.214) | (0.118) | (0.016) | (0.138) |
| 200 | 100 | 6.61 | 0.69 | 6.24 | 5.48 | 0.52 | 2.75 | **2.12** | **0.50** | **1.26** |
|    |     | (0.035) | (0.017) | (0.133) | (0.062) | (0.019) | (0.147) | (0.046) | (0.006) | (0.031) |

Table 4: Comparison of estimation performance of three methods with diagonal covariance matrix over 1,000 replications. The standard deviations are presented in the parentheses. Here $L_\mathsf{F}, L_2$, and $L_1$ represent the Frobenius, $L_2$, and $L_1$ matrix norms respectively. The pattern of the transition matrix is "random".

|     |     | ridge method | | | lasso method | | | our method | | |
| --- | --- | --- | --- | --- | --- | --- | --- | --- | --- | --- |
| $d$ | $T$ | $L_\mathsf{F}$ | $L_2$ | $L_1$ | $L_\mathsf{F}$ | $L_2$ | $L_1$ | $L_\mathsf{F}$ | $L_2$ | $L_1$ |
| 50 | 100 | 2.60 | 0.48 | 2.45 | 2.21 | **0.43** | 1.53 | **1.73** | 0.44 | **0.73** |
|    |     | (0.031) | (0.027) | (0.102) | (0.061) | (0.030) | (0.143) | (0.051) | (0.026) | (0.034) |
| 100 | 50 | 4.10 | 0.61 | 3.53 | 5.44 | 0.71 | 3.09 | **3.07** | **0.48** | **1.21** |
|    |     | (0.025) | (0.020) | (0.136) | (0.077) | (0.024) | (0.224) | (0.066) | (0.024) | (0.177) |
| 200 | 100 | 7.01 | 0.74 | 6.27 | 6.03 | 0.58 | 2.79 | **3.54** | **0.44** | **0.95** |
|    |     | (0.024) | (0.019) | (0.179) | (0.048) | (0.011) | (0.163) | (0.036) | (0.026) | (0.079) |

Table 5: Comparison of estimation performance of three methods with diagonal covariance matrix over 1,000 replications. The standard deviations are presented in the parentheses. Here $L_\mathsf{F}, L_2$, and $L_1$ represent the Frobenius, $L_2$, and $L_1$ matrix norms respectively. The pattern of the transition matrix is "scale-free".

|     |     | ridge method | | | lasso method | | | our method | | |
| --- | --- | --- | --- | --- | --- | --- | --- | --- | --- | --- |
| $d$ | $T$ | $L_\mathsf{F}$ | $L_2$ | $L_1$ | $L_\mathsf{F}$ | $L_2$ | $L_1$ | $L_\mathsf{F}$ | $L_2$ | $L_1$ |
| 50 | 100 | 2.48 | 0.44 | 2.40 | 2.09 | 0.41 | 1.51 | **1.44** | **0.41** | **0.98** |
|     |     | (0.032) | (0.025) | (0.098) | (0.059) | (0.033) | (0.154) | (0.075) | (0.052) | (0.108) |
| 100 | 50 | 3.60 | 0.56 | 3.43 | 5.14 | 0.64 | 3.11 | **2.16** | **0.46** | **1.36** |
|     |     | (0.034) | (0.023) | (0.133) | (0.085) | (0.031) | (0.188) | (0.130) | (0.043) | (0.115) |
| 200 | 100 | 6.65 | 0.70 | 6.26 | 5.57 | 0.51 | 3.29 | **2.51** | **0.42** | **2.49** |
|     |     | (0.034) | (0.017) | (0.143) | (0.065) | (0.014) | (0.274) | (0.249) | (0.050) | (0.108) |

Table 6: Comparison of estimation performance of three methods on data generated with Toeplitz covariance matrix ($\rho = 0.5$), over 1,000 replications. The standard deviations are presented in the parentheses. Here $L_\mathsf{F}, L_2$, and $L_1$ represent the Frobenius, $L_2$, and $L_1$ matrix norms respectively. The pattern of the transition matrix is "band".

|     |     | ridge method | | | lasso method | | | our method | | |
| --- | --- | --- | --- | --- | --- | --- | --- | --- | --- | --- |
| $d$ | $T$ | $L_\mathsf{F}$ | $L_2$ | $L_1$ | $L_\mathsf{F}$ | $L_2$ | $L_1$ | $L_\mathsf{F}$ | $L_2$ | $L_1$ |
| 50 | 100 | 2.47 | 0.51 | 2.25 | 2.10 | **0.45** | 1.32 | **1.82** | 0.47 | **0.57** |
|     |     | (0.031) | (0.033) | (0.101) | (0.066) | (0.035) | (0.131) | (0.084) | (0.014) | (0.044) |
| 100 | 50 | 3.98 | 0.67 | 3.31 | 5.22 | 0.74 | 2.81 | **3.15** | **0.51** | **1.04** |
|     |     | (0.029) | (0.033) | (0.107) | (0.083) | (0.032) | (0.174) | (0.114) | (0.063) | (0.529) |
| 200 | 100 | 6.92 | 0.79 | 5.96 | 5.82 | 0.61 | 2.44 | **3.79** | **0.48** | **0.67** |
|     |     | (0.033) | (0.028) | (0.142) | (0.060) | (0.023) | (0.134) | (0.078) | (0.006) | (0.034) |

Table 7: Comparison of estimation performance of three methods on data generated with Toeplitz covariance matrix ($\rho = 0.5$), over 1,000 replications. The standard deviations are presented in the parentheses. Here $L_\mathsf{F}, L_2$, and $L_1$ represent the Frobenius, $L_2$, and $L_1$ matrix norms respectively. The pattern of the transition matrix is "cluster".

|     |     | ridge method | | | lasso method | | | our method | | |
| --- | --- | --- | --- | --- | --- | --- | --- | --- | --- | --- |
| $d$ | $T$ | $L_\mathsf{F}$ | $L_2$ | $L_1$ | $L_\mathsf{F}$ | $L_2$ | $L_1$ | $L_\mathsf{F}$ | $L_2$ | $L_1$ |
| 50  | 100 | 2.32 | 0.42 | 2.25 | 2.01 | **0.39** | 1.42 | **1.46** | 0.47 | **0.69** |
|     |     | (0.041) | (0.029) | (0.114) | (0.066) | (0.030) | (0.124) | (0.027) | (0.019) | (0.037) |
| 100 | 50  | 3.61 | 0.57 | 3.33 | 5.08 | 0.65 | 3.01 | **2.47** | **0.47** | **1.02** |
|     |     | (0.034) | (0.029) | (0.124) | (0.087) | (0.031) | (0.212) | (0.075) | (0.031) | (0.155) |
| 200 | 100 | 6.63 | 0.70 | 6.13 | 5.58 | 0.54 | 2.59 | **2.96** | **0.48** | **0.79** |
|     |     | (0.038) | (0.020) | (0.162) | (0.069) | (0.019) | (0.153) | (0.027) | (0.013) | (0.046) |

Table 8: Comparison of estimation performance of three methods on data generated with Toeplitz covariance matrix ($\rho = 0.5$), over 1,000 replications. The standard deviations are presented in the parentheses. Here $L_\mathsf{F}, L_2$, and $L_1$ represent the Frobenius, $L_2$, and $L_1$ matrix norms respectively. The pattern of the transition matrix is "hub".

|     |     | ridge method | | | lasso method | | | our method | | |
| --- | --- | --- | --- | --- | --- | --- | --- | --- | --- | --- |
| $d$ | $T$ | $L_\mathsf{F}$ | $L_2$ | $L_1$ | $L_\mathsf{F}$ | $L_2$ | $L_1$ | $L_\mathsf{F}$ | $L_2$ | $L_1$ |
| 50  | 100 | 2.27 | 0.40 | 2.22 | 1.85 | **0.36** | 1.34 | **1.16** | 0.39 | **1.01** |
|     |     | (0.039) | (0.037) | (0.099) | (0.067) | (0.041) | (0.157) | (0.124) | (0.062) | (0.102) |
| 100 | 50  | 3.37 | 0.54 | 3.26 | 4.94 | 0.61 | 2.96 | **1.86** | **0.50** | **1.37** |
|     |     | (0.041) | (0.034) | (0.125) | (0.102) | (0.033) | (0.222) | (0.120) | (0.017) | (0.104) |
| 200 | 100 | 6.46 | 0.67 | 6.19 | 5.24 | 0.50 | 2.54 | **2.13** | **0.49** | **1.24** |
|     |     | (0.042) | (0.024) | (0.168) | (0.071) | (0.025) | (0.162) | (0.107) | (0.023) | (0.042) |

Table 9: Comparison of estimation performance of three methods on data generated with Toeplitz covariance matrix ($\rho = 0.5$), over 1,000 replications. The standard deviations are presented in the parentheses. Here $L_\mathsf{F}, L_2$, and $L_1$ represent the Frobenius, $L_2$, and $L_1$ matrix norms respectively. The pattern of the transition matrix is "random".

|   |   | ridge method | | | lasso method | | | our method | | |
|---|---|---|---|---|---|---|---|---|---|---|
| $d$ | $T$ | $L_\mathsf{F}$ | $L_2$ | $L_1$ | $L_\mathsf{F}$ | $L_2$ | $L_1$ | $L_\mathsf{F}$ | $L_2$ | $L_1$ |
| 50 | 100 | 2.49 | 0.45 | 2.34 | 2.15 | **0.41** | 1.44 | **1.74** | 0.44 | **0.74** |
|   |   | (0.036) | (0.029) | (0.104) | (0.071) | (0.032) | (0.139) | (0.058) | (0.033) | (0.043) |
| 100 | 50 | 4.02 | 0.60 | 3.42 | 5.34 | 0.70 | 2.96 | **3.07** | **0.47** | **1.21** |
|   |   | (0.029) | (0.024) | (0.123) | (0.092) | (0.028) | (0.207) | (0.085) | (0.027) | (0.192) |
| 200 | 100 | 6.89 | 0.72 | 6.13 | 5.87 | 0.56 | 2.65 | **3.54** | **0.43** | **0.97** |
|   |   | (0.028) | (0.022) | (0.164) | (0.057) | (0.016) | (0.174) | (0.052) | (0.019) | (0.091) |

Table 10: Comparison of estimation performance of three methods on data generated with Toeplitz covariance matrix ($\rho = 0.5$), over 1,000 replications. The standard deviations are presented in the parentheses. Here $L_\mathsf{F}, L_2$, and $L_1$ represent the Frobenius, $L_2$, and $L_1$ matrix norms respectively. The pattern of the transition matrix is "scale-free".

|   |   | ridge method | | | lasso method | | | our method | | |
|---|---|---|---|---|---|---|---|---|---|---|
| $d$ | $T$ | $L_\mathsf{F}$ | $L_2$ | $L_1$ | $L_\mathsf{F}$ | $L_2$ | $L_1$ | $L_\mathsf{F}$ | $L_2$ | $L_1$ |
| 50 | 100 | 2.36 | 0.42 | 2.27 | 2.00 | 0.38 | 1.36 | **1.42** | **0.37** | **0.89** |
|   |   | (0.036) | (0.033) | (0.094) | (0.064) | (0.033) | (0.136) | (0.068) | (0.056) | (0.108) |
| 100 | 50 | 3.49 | 0.55 | 3.29 | 5.03 | 0.63 | 2.96 | **2.21** | **0.42** | **1.29** |
|   |   | (0.039) | (0.029) | (0.124) | (0.100) | (0.027) | (0.212) | (0.149) | (0.050) | (0.131) |
| 200 | 100 | 6.52 | 0.67 | 6.18 | 5.36 | 0.49 | 3.06 | **2.55** | **0.39** | **2.44** |
|   |   | (0.041) | (0.019) | (0.165) | (0.070) | (0.013) | (0.219) | (0.364) | (0.062) | (0.134) |

Table 11: Comparison of estimation performance of three methods over 1,000 replications under multiple lag settings. The standard deviations are presented in the parentheses. Here $L_{\mathsf{F}}, L_2$, and $L_1$ represent the Frobenius, $L_2$, and $L_1$ matrix norms respectively. The pattern of the transition matrix is "hub".

|   | ridge method | | | lasso method | | | our method | | |
|---|---|---|---|---|---|---|---|---|---|
| $p$ | $L_{\mathsf{F}}$ | $L_2$ | $L_1$ | $L_{\mathsf{F}}$ | $L_2$ | $L_1$ | $L_{\mathsf{F}}$ | $L_2$ | $L_1$ |
| 1 | 6.93 | 2.50 | 7.35 | 1.83 | 0.52 | 1.36 | **0.25** | **0.11** | **0.23** |
|   | (0.012) | (0.094) | (0.377) | (0.039) | (0.017) | (0.128) | (0.014) | (0.016) | (0.002) |
| 3 | 9.13 | 2.89 | 15.96 | 2.52 | 0.59 | 2.18 | **0.45** | **0.18** | **0.70** |
|   | (0.129) | (0.092) | (0.249) | (0.085) | (0.016) | (0.116) | (0.023) | (0.004) | (0.003) |
| 5 | 5.57 | 1.57 | 11.73 | 2.75 | 0.61 | 3.19 | **0.58** | **0.23** | **1.23** |
|   | (0.000) | (0.000) | (0.000) | (0.000) | (0.000) | (0.000) | (0.000) | (0.000) | (0.000) |
| 7 | 4.27 | 1.14 | 10.92 | 2.90 | 0.60 | 3.44 | **0.72** | **0.31** | **1.83** |
|   | (0.010) | (0.041) | (0.152) | (0.026) | (0.025) | (0.183) | (0.077) | (0.067) | (0.222) |
| 9 | 3.59 | 0.90 | 10.17 | 2.98 | 0.61 | 4.11 | **0.70** | **0.30** | **2.11** |
|   | (0.026) | (0.023) | (0.219) | (0.061) | (0.004) | (0.201) | (0.000) | (0.000) | (0.000) |

With $\{A_i\}_{i=1}^p, \widetilde{\Sigma}$, and $\Psi$ determined, we simulate a time series $(X_1, \ldots, X_T)^{\mathrm{T}} \in \mathbb{R}^{T \times d}$ according to the model described in (2.1) (with lag $p \geq 1$).

The estimation error is calculated by measuring the difference of $(A_1^{\mathrm{T}}, \ldots, A_p^{\mathrm{T}})^{\mathrm{T}}$ and $(\widehat{A}_1^{\mathrm{T}}, \ldots, \widehat{A}_p^{\mathrm{T}})^{\mathrm{T}}$ with regard to different matrix norms ($L_{\mathsf{F}}, L_2$, and $L_1$ norms). We conduct 1,000 simulations and compare the averaged performance of three competing methods. The calculated averaged estimation errors are illustrated in Tables 11 and 12. The standard deviations of the estimation errors are provided in the parentheses. Here the tuning parameters are selected in the same way as before. Tables 11 and 12 confirms that our method still outperforms the competing two methods.

### 5.2.3 Synthetic Data: Impact of Transition Matrices' Spectral Norms

In this section we illustrate the effects of the transition matrices' spectral norms on estimation accuracy. To this end, we study the settings in Section 5.2. More specifically, we set lag $p = 1$, the dimension $d$ and the sample size $T$ to be $d = 50$ and $T = 100$. The transition matrix $A_1$ is created according to different patterns ("band", "cluster", "hub", "scale-free", and "random"), and then rescaled such that $\|A_1\|_2 = \kappa$, where $\kappa$ is from 0.05 to 0.9. Covariance matrix $\Sigma$ is set to be of the form (5.1), and $\Psi$ is accordingly determined by stationary condition. We select the tuning parameters using the cross-validation procedure as before. The estimation errors are then plotted against $\kappa$ and shown in Figure 2.

Figure 2 illustrates that the estimation error is an increasing function of the spectral norm $\|A_1\|_2$. This demonstrates that the spectral norms of the transition matrices play an important role in estimation accuracy and justifies the theorems in Section 4.

Table 12: Comparison of estimation performance of three methods over 1,000 replications under multiple lag settings. The standard deviations are presented in the parentheses. Here $L_\mathsf{F}, L_2,$ and $L_1$ represent the Frobenius, $L_2$, and $L_1$ matrix norms respectively. The pattern of the transition matrix is "scale-free".

| | ridge method | | | lasso method | | | our method | | |
|---|---|---|---|---|---|---|---|---|---|
| $p$ | $L_\mathsf{F}$ | $L_2$ | $L_1$ | $L_\mathsf{F}$ | $L_2$ | $L_1$ | $L_\mathsf{F}$ | $L_2$ | $L_1$ |
| 1 | 6.93 | 2.51 | 7.39 | 1.83 | 0.53 | 1.35 | **0.30** | **0.12** | **0.24** |
| | (0.116) | (0.093) | (0.340) | (0.041) | (0.018) | (0.129) | (0.045) | (0.016) | (0.039) |
| 3 | 9.14 | 3.00 | 15.97 | 2.53 | 0.60 | 2.19 | **0.46** | **0.17** | **0.57** |
| | (0.133) | (0.099) | (0.219) | (0.090) | (0.020) | (0.094) | (0.058) | (0.007) | (0.083) |
| 5 | 5.58 | 1.57 | 11.66 | 2.77 | 0.60 | 2.97 | **0.62** | **0.23** | **0.93** |
| | (0.002) | (0.002) | (0.018) | (0.001) | (0.002) | (0.076) | (0.012) | (0.002) | (0.078) |
| 7 | 4.28 | 1.14 | 10.97 | 2.90 | 0.60 | 3.34 | **0.69** | **0.24** | **1.29** |
| | (0.014) | (0.042) | (0.164) | (0.031) | (0.020) | (0.131) | (0.041) | (0.005) | (0.078) |
| 9 | 3.62 | 0.90 | 10.25 | 3.01 | 0.61 | 3.42 | **0.87** | **0.30** | **1.79** |
| | (0.024) | (0.023) | (0.267) | (0.058) | (0.003) | (0.112) | (0.078) | (0.012) | (0.198) |

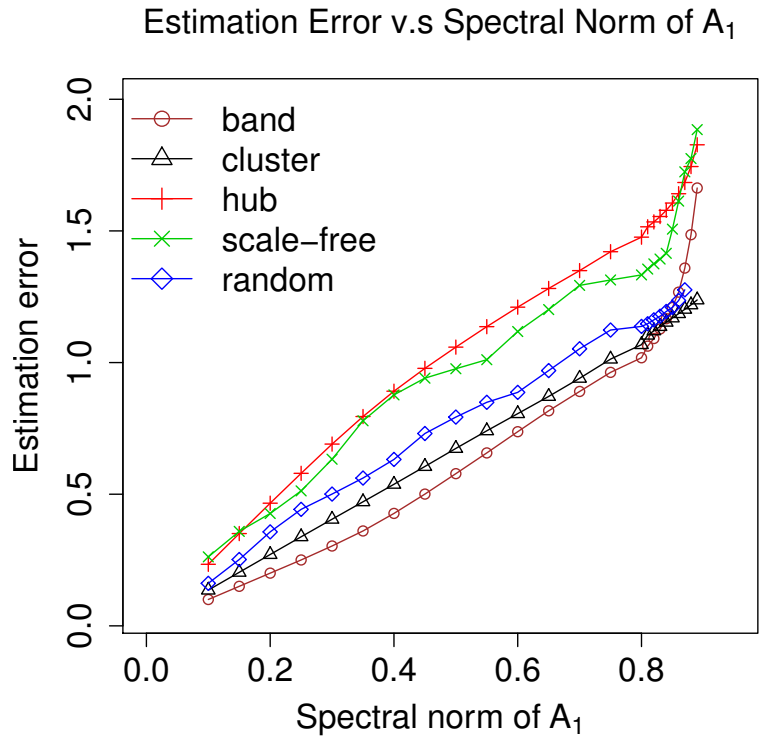

Figure 2: Estimation errors of $A_1$ (in $L_1$ norm) plotted against spectral norms of $A_1$.

Table 13: Memory usage v.s lag of model: The result shown below is the memory usage (in Mb) of a single replicate of experiment, with the lag $p$ changing from 1 to 9. The pattern of the transition matrices $\{A_i\}_{i=1}^{p}$ is "random".

| Lag of model (p) | 1 | 2 | 3 | 4 | 5 | 6 | 7 | 8 | 9 |
|---|---|---|---|---|---|---|---|---|---|
| Mem. Use (Mb) | 5.566 | 8.724 | 11.862 | 14.999 | 18.135 | 21.272 | 24.406 | 27.540 | 30.673 |

### 5.2.4 Computation Time and Memory Usage

This section is devoted to show the computation time and memory usage of our method. First, we show the advantage of our method in terms of computation time. A major advantage of our method over the two competing methods is that our method can be easily parallelly computed and thus has the potential to save computation time. We illustrate this point with a plot of the computation time as a function of the number of available cores. All experiments are conducted on an 2816-core Westmere/Ivybridge 2.67/2.5GHz Linux server with 17T memory, a cluster system with batch scheduling.

To this end, we set the time series length $T = 100$ and the dimension $d = 50$. The transition matrix $A_1$ is created according to the pattern "random", and then rescaled such that $\|A_1\|_2 = 0.5$. The covariance matrix $\Sigma$ is generated as in (5.1), and $\Psi$ is generated by stationary condition. We then solve (3.5) using parallel computation based on 1 to 50 cores.

Figure 3 shows the computation time. It illustrates that, in terms of saving computation time, under this specific setting, we have: (i) Our method outperforms the ridge method even if we do not parallelly compute it; (ii) When there are no less than 16 cores, our method outperforms the lasso method. Here the ridge method is very slow because it involves calculating the inverse of a large matrix.

Secondly, we show the memory usage of our method. By converting the time series from VAR(1) to VAR(p), the memory usage increases. For investigating the memory usage, we conduct an empirical study. Specifically, we choose the lag $p$ to be $1, 2, \ldots, 9$, the time series length $T = 100$, and the dimension $d = 50$. Transition matrices $A_1, \ldots, A_p$ are created according to the "random" pattern, and then rescaled such that $\|A_i\|_2 = 0.1$ for $i = 1, \ldots, p$. $\Psi$ is set as $I_d$ for simplicity. With $\{A_i\}_{i=1}^{p}$ and $\Psi$, we simulate a time series $(X_1, \ldots, X_T)^T \in \mathbb{R}^{T \times d}$ according to (2.1) with lag $p \geq 1$. The result in Table 13 is the memory usage of a single replicate of experiment in megabytes (Mb). It shows that the memory usage is approximately increasing linearly with regard to $p$ under this setting.

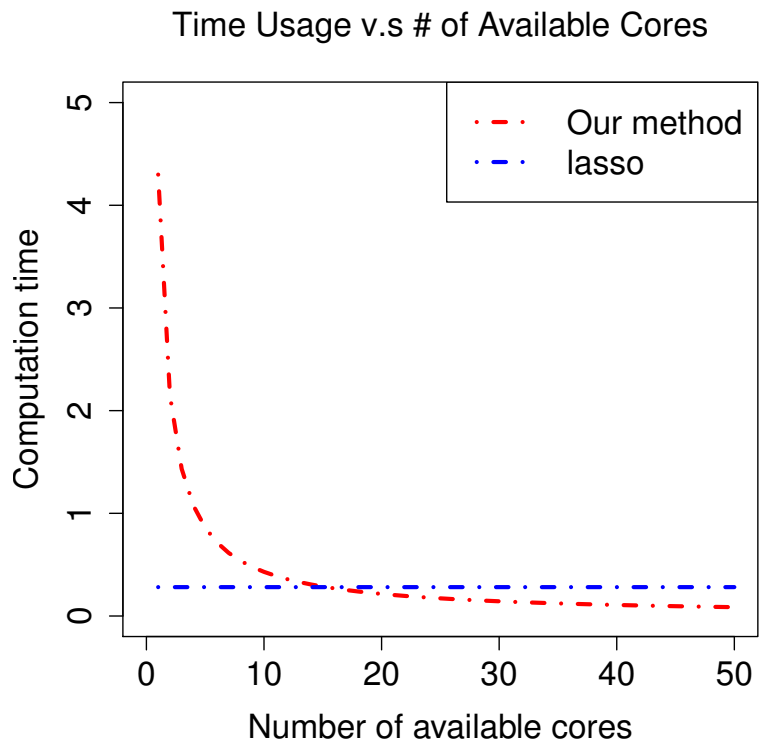

Figure 3: Computation time v.s number of available cores. The computation time for ridge and lasso are 5.593s and 0.281s, which do not change with number of available cores. The computation time here is the elapsed time (in seconds) of a single replicate of experiment.

## 5.3 Real Data

We further compare the three methods on the equity data collected from Yahoo! Finance. The task is to predict the stock prices. We collect the daily closing prices for 91 stocks that are consistently in the S&P 100 index between January 1, 2003 and January 1, 2008. This gives us altogether 1,258 data points, each of which corresponds to the vector of closing prices on a trading day.

We first provide comparison on averaged prediction errors for using different lag $p$ on this dataset. Let $E = (E_{t,j}) \in \mathbb{R}^{1258 \times 91}$ with $E_{t,j}$ denoting the closing price of the stock $j$ on day $t$. We screen out all the stocks with low marginal standard deviations and only keep 50 stocks which vary the most. We center the data so that the marginal mean of each time series is zero. The resulting data matrix is denoted by $\bar{E} \in \mathbb{R}^{1258 \times 50}$. We apply the three methods on $\bar{E}$ with different lag $p$ changing from 1 to 9. To evaluate the performance of the three methods, for $t = 1248, \ldots, 1257$, we select the dataset $\bar{E}_{J_t, *}$, where we have $J_t = \{j : t - 100 \leq j \leq t - 1\}$, as the training set. Then for each $p$ and $\lambda$, based on the training set $\bar{E}_{J_t, *}$, we calculate the transition matrix estimates $\widehat{A}_1^t(p, \lambda), \ldots, \widehat{A}_p^t(p, \lambda)$. We then use the obtained estimates to predict the stock price in day $t$. The averaged prediction error for each specific $\lambda$ and $p$ is calculated as

$$\overline{\mathrm{Err}}(p, \lambda) = \frac{1}{10} \sum_{t=1}^{10} \|\bar{E}_{t,*} - \sum_{k=1}^{p} \widehat{A}_k^t(p, \lambda)^{\mathrm{T}} \bar{E}_{t-k,*}\|_2.$$

In Table 14, we present the minimized averaged prediction errors $\min_\lambda \overline{\mathrm{Err}}(p, \lambda)$ for the three methods with different lag $p$. The standard deviations of the prediction errors are presented in the parentheses. Our method outperforms the two competing methods in terms of prediction accuracy.

Secondly, we provide the prediction error on day $t = 1258$ based on the selected $(p, \lambda)$ using cross-validation. By observing Table 14, we select the lag $p = 1$ and the corresponding $\lambda$ for our method. The prediction error is 7.62 for our method. In comparison, the lasso method and ridge method have the prediction errors 11.11 and 11.94 separately.

## 6 Discussions

Estimation of the vector autoregressive model is an interesting problem and has been investigated for a long time. This problem is intrinsically linked to the regression problem with multiple responses. Accordingly (penalized) least squares estimates, which has the maximum likelihood interpretation behind it, look like reasonable solutions. However, high dimensionality brings significantly new challenges and viewpoints to this classic problem. In parallel to the Dantzig selector proposed by Candes and Tao (2007) in cracking the ordinary linear regression model, we advocate borrowing the strength of the linear program in estimating the VAR model. As has been repeatedly stated in the main text, this new formulation brings some advantages over the least square estimates. Moreover, our theoretical analysis brings new insights into the problem of transition matrix estimation, and we highlight the role of $\|A_1\|_2$ in evaluating the estimation accuracy of the estimator.

Table 14: The optimized averaged prediction errors for the three methods on the equity data, under different lags $p$ from 1 to 9. The standard deviations are present in the parentheses.

| lag | ridge method | lasso method | our method |
| --- | --- | --- | --- |
| $p=1$ | 17.68 (2.49) | 15.67 (2.74) | **11.88** (3.34) |
| $p=2$ | 15.63 (3.01) | 15.69 (2.84) | 12.01 (3.41) |
| $p=3$ | 15.17 (3.53) | 15.76 (2.83) | 12.04 (3.42) |
| $p=4$ | 14.90 (3.69) | 15.68 (2.76) | 12.02 (3.41) |
| $p=5$ | 14.73 (3.66) | 15.62 (2.55) | 12.08 (3.29) |
| $p=6$ | 14.58 (3.57) | 15.51 (2.58) | 12.09 (3.15) |
| $p=7$ | 14.42 (3.49) | 15.45 (2.59) | 12.21 (3.16) |
| $p=8$ | 14.36 (3.42) | 15.40 (2.57) | 12.25 (3.16) |
| $p=9$ | **14.20** (3.31) | **15.28** (2.46) | 12.24 (3.06) |

In the main text we do not discuss estimating the covariance matrix $\Sigma$ and $\Psi$. Lemma A.1 builds the $L_{\max}$ convergence result for estimating $\Sigma$. If we further suppose that the covariance matrix $\Sigma$ is sparse in some sense, then we can exploit the well developed results in covariance matrix estimation (including "banding" (Bickel and Levina, 2008b), "tapering" (Cai et al., 2010), and "thresholding" (Bickel and Levina, 2008a)) to estimate the covariance matrix $\Sigma$ and establish the consistency result with regard to the matrix $L_1$ and $L_2$ norms. With both $\Sigma$ and $A$ estimated by some constant estimator $\widehat{\Sigma}$, an estimator $\widehat{\Psi}$ of $\Psi$ can be obtained under the VAR model (with lag one) as:

$$\widehat{\Psi} = \widehat{\Sigma} - \widehat{A}_1^{\mathrm{T}} \widehat{\Sigma} \widehat{A}_1,$$

and a similar estimator can be built for lag $p$ VAR model using the augmented formulation shown in Equation (2.3).

In this manuscript we focus on the stationary vector autoregressive model and our method is designed for such stationary process. The stationary requirement is a common assumption in analysis and is adopted by most recent works, for example, Kock and Callot (2012) and Song and Bickel (2011). We notice that there are works in handling unstable VAR models, checking for example Song et al. (2014) and Kock (2012). We would like to explore this problem in the future. Another unexplored region is how to determine the order (lag) of the vector autoregression aside from using the cross-validation approach. There have been results in this area (e.g., Song and Bickel (2011)) and we are also interested in finding whether the linear program can also be exploited in determining the order of the VAR model.

# A  Proofs of Main Results

In this section we provide the proofs of the main results in the manuscript.

## A.1  Proof of Theorem 4.1

Before proving the main result in Theorem 4.1, we first establish several lemmas. In the sequel, because we only focus on the lag 1 autoregressive model, for notation simplicity, in $\Sigma_i(\{(X_t)\})$ we remove $\{(X_t)\}$ and simply denote the lag $i$ covariance matrix to be $\Sigma_i$.

The following lemma describes the $L_{\max}$ rate of convergence $S$ to $\Sigma$. This result generalizes the upper bound derived when data are independently generated (see, for example, Bickel and Levina (2008a)).

**Lemma A.1.** *Letting $S$ be the marginal sample covariance matrix defined in (3.1), when $T \geq \max(6 \log d, 1)$, we have, with probability no smaller than $1 - 6d^{-1}$,*

$$\|S - \Sigma\|_{\max} \leq \frac{16\|\Sigma\|_2 \max_j(\Sigma_{jj})}{\min_j(\Sigma_{jj})(1 - \|A_1\|_2)} \left\{ \left(\frac{6 \log d}{T}\right)^{1/2} + 2\left(\frac{1}{T}\right)^{1/2} \right\}.$$

*Proof.* For any $j, k \in \{1, 2, \ldots, d\}$, we have

$$\mathbb{P}(|S_{jk} - \Sigma_{jk}| > \eta) = \mathbb{P}\left(\left|\frac{1}{T}\sum_{t=1}^T X_{tj} X_{tk} - \Sigma_{jk}\right| > \eta\right).$$

Letting $Y_t = \{X_{t1}(\Sigma_{11})^{-1/2}, \ldots, X_{td}(\Sigma_{dd})^{-1/2}\}^{\mathrm{T}}$ for $t = 1, \ldots, T$ and $\rho_{jk} = \Sigma_{jk}(\Sigma_{jj}\Sigma_{kk})^{-1/2}$, we have

$$\begin{aligned}
\mathbb{P}(|S_{jk} - \Sigma_{jk}| > \eta) &= \mathbb{P}\left\{\left|\frac{1}{T}\sum_{t=1}^T Y_{tj} Y_{tk} - \rho_{jk}\right| > \eta(\Sigma_{jj}\Sigma_{kk})^{-1/2}\right\} \\
&= \mathbb{P}\left\{\left|\frac{\sum_{t=1}^T (Y_{tj} + Y_{tk})^2 - \sum_{t=1}^T (Y_{tj} - Y_{tk})^2}{4T} - \rho_{jk}\right| > \eta(\Sigma_{jj}\Sigma_{kk})^{-1/2}\right\} \\
&\leq \mathbb{P}\left\{\left|\frac{1}{T}\sum_{t=1}^T (Y_{tj} + Y_{tk})^2 - 2(1 + \rho_{jk})\right| > 2\eta(\Sigma_{jj}\Sigma_{kk})^{-1/2}\right\} \\
&\quad + \mathbb{P}\left\{\left|\frac{1}{T}\sum_{t=1}^T (Y_{tj} - Y_{tk})^2 - 2(1 - \rho_{jk})\right| > 2\eta(\Sigma_{jj}\Sigma_{kk})^{-1/2}\right\}. \quad (A.1)
\end{aligned}$$

Using the property of Gaussian distribution, we have $(Y_{1j}+Y_{1k}, \ldots, Y_{Tj}+Y_{Tk})^{\mathrm{T}} \sim N_T(0, Q)$ for some positive definite matrix $Q$. In particular, we have

$$\begin{aligned}
|Q_{il}| &= |\operatorname{Cov}(Y_{ij}+Y_{ik}, Y_{lj}+Y_{lk})| = |\operatorname{Cov}(Y_{ij}, Y_{lj}) + \operatorname{Cov}(Y_{ij}, Y_{lk}) + \operatorname{Cov}(Y_{ik}, Y_{lk}) + \operatorname{Cov}(Y_{ik}, Y_{lj})| \\
&\leq \frac{1}{\min_j(\Sigma_{jj})}|\operatorname{Cov}(X_{ij}, X_{lj}) + \operatorname{Cov}(X_{ij}, X_{lk}) + \operatorname{Cov}(X_{ik}, X_{lk}) + \operatorname{Cov}(X_{ik}, X_{lj})| \\
&\leq \frac{4}{\min_j(\Sigma_{jj})}\|\Sigma_{l-i}\|_{\max} \leq \frac{8\|\Sigma\|_2 \|A_1\|_2^{|l-i|}}{\min_j(\Sigma_{jj})},
\end{aligned}$$

where the last inequality follows from (2.2).

Therefore, using the matrix norm inequality,

$$\|Q\|_2 \le \max_{1\le i\le T}\sum_{l=1}^{T}|Q_{il}| \le \frac{8\|\Sigma\|_2}{\min_j(\Sigma_{jj})(1-\|A_1\|_2)}.$$

Then applying Lemma B.1 to (A.1), we have

$$\mathbb{P}\left\{\left|\frac{1}{T}\sum_{t=1}^{T}(Y_{tj}+Y_{tk})^2 - 2(1+\rho_{jk})\right| > 2\eta(\Sigma_{jj}\Sigma_{kk})^{-1/2}\right\}$$
$$\le 2\exp\left[-\frac{T}{2}\left\{\frac{\eta\min_j(\Sigma_{jj})(1-\|A_1\|_2)}{16\|\Sigma\|_2(\Sigma_{jj}\Sigma_{kk})^{1/2}} - 2T^{-1/2}\right\}^2\right] + 2\exp\left(-\frac{T}{2}\right). \quad (A.2)$$

Using a similar argument, we have

$$\mathbb{P}\left\{\left|\frac{1}{T}\sum_{t=1}^{T}(Y_{tj}-Y_{tk})^2 - 2(1-\rho_{jk})\right| > 2\eta(\Sigma_{jj}\Sigma_{kk})^{-1/2}\right\}$$
$$\le 2\exp\left[-\frac{T}{2}\left\{\frac{\eta\min_j(\Sigma_{jj})(1-\|A_1\|_2)}{16\|\Sigma\|_2(\Sigma_{jj}\Sigma_{kk})^{1/2}} - 2T^{-1/2}\right\}^2\right] + 2\exp\left(-\frac{T}{2}\right). \quad (A.3)$$

Combining (A.2) and (A.3), then applying the union bound, we have

$$\mathbb{P}(\|S-\Sigma\|_{\max}>\eta)$$
$$\le 3d^2\exp\left(-\frac{T}{2}\right) + 3d^2\exp\left[-\frac{T}{2}\left\{\frac{\eta\min_j(\Sigma_{jj})(1-\|A_1\|_2)}{16\|\Sigma\|_2\max_j(\Sigma_{jj})} - 2\left(\frac{1}{T}\right)^{-1/2}\right\}^2\right].$$

The proof thus completes by choosing $\eta$ as the described form. $\square$

In the next lemma we try to quantify the difference between $S_1$ and $\Sigma_1$ with respect to the matrix $L_{\max}$ norm. Remind that $\Sigma_1\{(X_t)\}$ is simplified to be $\Sigma_1$.

**Lemma A.2.** *Letting $S_1$ be the lag 1 sample covariance matrix, when $T \ge \max(6\log d + 1, 2)$, we have, with probability no smaller than $1 - 8d^{-1}$,*

$$\|S_1 - \Sigma_1\|_{\max} \le \frac{32\|\Sigma\|_2\max_j(\Sigma_{jj})}{\min_j(\Sigma_{jj})(1-\|A_1\|_2)}\left\{\left(\frac{3\log d}{T}\right)^{1/2} + \left(\frac{2}{T}\right)^{1/2}\right\}.$$

*Proof.* We have, for any $j,k \in \{1,2,\ldots,d\}$,

$$\mathbb{P}(|(S_1)_{jk} - (\Sigma_1)_{jk}| > \eta) = \mathbb{P}\left(\left|\frac{1}{T-1}\sum_{t=1}^{T-1}X_{tj}X_{(t+1)k} - (\Sigma_1)_{jk}\right| > \eta\right).$$

Letting $Y_t = \{X_{t1}(\Sigma_{11})^{-1/2}, \ldots, X_{td}(\Sigma_{dd})^{-1/2}\}^{\mathrm{T}}$ and $\rho_{jk} = (\Sigma_1)_{jk}(\Sigma_{jj}\Sigma_{kk})^{-1/2}$, we have

$$\mathbb{P}(|(S_1)_{jk} - (\Sigma_1)_{jk}| > \eta) = \mathbb{P}\left\{\left|\frac{1}{T-1}\sum_{t=1}^{T-1} Y_{tj}Y_{(t+1)k} - \rho_{jk}\right| > \eta(\Sigma_{jj}\Sigma_{kk})^{-1/2}\right\}$$

$$= \mathbb{P}\left[\left|\frac{\sum_{t=1}^{T-1}\{Y_{tj}+Y_{(t+1)k}\}^2 - \sum_{t=1}^{T-1}\{Y_{tj}-Y_{(t+1)k}\}^2}{4(T-1)} - \rho_{jk}\right| > \eta(\Sigma_{jj}\Sigma_{kk})^{-1/2}\right]$$

$$\leq \mathbb{P}\left[\left|\frac{\sum_{t=1}^{T-1}\{Y_{tj}+Y_{(t+1)k}\}^2}{T-1} - 2(1+\rho_{jk})\right| > 2\eta(\Sigma_{jj}\Sigma_{kk})^{-1/2}\right]$$

$$+ \mathbb{P}\left[\left|\frac{\sum_{t=1}^{T-1}\{Y_{tj}-Y_{(t+1)k}\}^2}{T-1} - 2(1-\rho_{jk})\right| > 2\eta(\Sigma_{jj}\Sigma_{kk})^{-1/2}\right]. \quad (A.4)$$

Using the property of Gaussian distribution, we have $\{Y_{1j}+Y_{2k}, \ldots, Y_{(T-1)j}+Y_{Tk}\}^{\mathrm{T}} \sim N_{T-1}(0, Q)$, for some positive definite matrix $Q$. In particular, we have

$$|Q_{il}| = |\operatorname{Cov}\{Y_{ij}+Y_{(i+1)k}, Y_{lj}+Y_{(l+1)k}\}|$$
$$= |\operatorname{Cov}(Y_{ij}, Y_{lj}) + \operatorname{Cov}\{Y_{ij}, Y_{(l+1)k}\} + \operatorname{Cov}\{Y_{(i+1)k}, Y_{lj}\} + \operatorname{Cov}\{Y_{(i+1)k}, Y_{(l+1)k}\}|$$
$$\leq \frac{1}{\min_j(\Sigma_{jj})}|\operatorname{Cov}(X_{ij}, X_{lj}) + \operatorname{Cov}\{X_{ij}, X_{(l+1)k}\} + \operatorname{Cov}\{X_{(i+1)k}, X_{lj}\} + \operatorname{Cov}\{X_{(i+1)k}, X_{(l+1)j}\}|$$
$$\leq \frac{2\|\Sigma_{l-i}\|_{\max} + \|\Sigma_{l+1-i}\|_{\max} + \|\Sigma_{l-1-i}\|_{\max}}{\min_j(\Sigma_{jj})}$$
$$\leq \frac{\|\Sigma\|_2(2\|A_1\|_2^{|l-i|} + \|A_1\|_2^{|l+1-i|} + \|A_1\|_2^{|l-1-i|})}{\min_j(\Sigma_{jj})}.$$

Therefore, using the matrix norm inequality,

$$\|Q\|_2 \leq \max_{1 \leq i \leq (T-1)} \sum_{l=1}^{T-1} |Q_{il}| \leq \frac{8\|\Sigma\|_2}{\min_j(\Sigma_{jj})(1-\|A_1\|_2)}.$$

Then applying Lemma B.1 to (A.4), we have

$$\mathbb{P}\left[\left|\frac{1}{T-1}\sum_{t=1}^{T-1}\{Y_{tj}+Y_{(t+1)k}\}^2 - 2(1+\rho_{jk})\right| > 2\eta(\Sigma_{jj}\Sigma_{kk})^{-1/2}\right] \leq$$

$$2\exp\left[-\frac{(T-1)}{2}\left\{\frac{\eta\min_j(\Sigma_{jj})(1-\|A_1\|_2)}{16\|\Sigma\|_2(\Sigma_{jj}\Sigma_{kk})^{1/2}} - 2(T-1)^{-1/2}\right\}^2\right] + 2\exp\left(-\frac{T-1}{2}\right). \quad (A.5)$$

Using a similar technique, we have

$$\mathbb{P}\left[\left|\frac{1}{T-1}\sum_{t=1}^{T-1}\{Y_{tj}-Y_{(t+1)k}\}^2 - 2(1-\rho_{jk})\right| > 2\eta(\Sigma_{jj}\Sigma_{kk})^{-1/2}\right] \leq$$

$$2\exp\left[-\frac{(T-1)}{2}\left\{\frac{\eta\min_j(\Sigma_{jj})(1-\|A_1\|_2)}{16\|\Sigma\|_2(\Sigma_{jj}\Sigma_{kk})^{1/2}} - 2(T-1)^{-1/2}\right\}^2\right] + 2\exp\left(-\frac{T-1}{2}\right). \quad (A.6)$$

Combining (A.5) and (A.6), and applying the union bound across all pairs $(j,k)$, we have

$$\mathbb{P}(\|S_1 - \Sigma_1\|_{\max} > \eta) \le$$
$$4d^2 \exp\left[-\frac{(T-1)}{2}\left\{\frac{\eta \min_j(\Sigma_{jj})(1 - \|A_1\|_2)}{16\|\Sigma\|_2 \max_j(\Sigma_{jj})} - 2(T-1)^{-1/2}\right\}^2\right] + 4d^2 \exp\left(-\frac{T-1}{2}\right). \quad (A.7)$$

Finally noting that when $T \ge 3$, we have $1/(T-1) < 2/T$. The proof thus completes by choosing $\eta$ as stated. $\square$

Using the above two technical lemmas, we can then proceed to the proof of the main results in Theorem 4.1.

*Proof of Theorem 4.1.* With Lemmas A.1 and A.2, we proceed to prove Theorem 4.1. We first denote

$$\zeta_1 = \frac{16\|\Sigma\|_2 \max_j(\Sigma_{jj})}{\min_j(\Sigma_{jj})(1 - \|A_1\|_2)}\left\{\left(\frac{6 \log d}{T}\right)^{1/2} + 2\left(\frac{1}{T}\right)^{1/2}\right\},$$

$$\zeta_2 = \frac{32\|\Sigma\|_2 \max_j(\Sigma_{jj})}{\min_j(\Sigma_{jj})(1 - \|A_1\|)_2}\left\{\left(\frac{3 \log d}{T}\right)^{1/2} + \left(\frac{2}{T}\right)^{1/2}\right\}.$$

Using Lemmas A.1 and A.2, we have, with probability no smaller than $1 - 14d^{-1}$,

$$\|S - \Sigma\|_{\max} \le \zeta_1, \quad \|S_1 - \Sigma_1\|_{\max} \le \zeta_2.$$

We firstly prove that population quantity $A_1$ is a feasible solution to the optimization problem in (3.4) with probability no smaller than $1 - 14d^{-1}$:

$$\begin{aligned}
\|SA_1 - S_1\|_{\max} &= \|S\Sigma^{-1}\Sigma_1 - S_1\|_{\max} \\
&= \|S\Sigma^{-1}\Sigma_1^T - \Sigma_1 + \Sigma_1 - S_1\|_{\max} \\
&\le \|(S\Sigma^{-1} - I_d)\Sigma_1\|_{\max} + \|\Sigma_1 - S_1\|_{\max} \\
&\le \|(S - \Sigma)\Sigma^{-1}\Sigma_1\|_{\max} + \zeta_2 \\
&\le \zeta_1 \|A_1\|_1 + \zeta_2 \\
&\le \lambda_0.
\end{aligned}$$

The last inequality holds by using the condition that $d \ge 8$ implies that $1/T \le \log d/(2T)$. Therefore, $A_1$ is feasible in the optimization equation, by checking the equivalence between (3.4) and (3.5), we have $\|\widehat{\Omega}\|_1 \le \|A_1\|_1$ with probability no smaller than $1 - 14d^{-1}$. We

then have

$$
\begin{aligned}
\|\widehat{\Omega} - A_1\|_{\max} &= \|\widehat{\Omega} - \Sigma^{-1}\Sigma_1\|_{\max} \\
&= \|\Sigma^{-1}(\Sigma\widehat{\Omega} - \Sigma_1)\|_{\max} \\
&= \|\Sigma^{-1}(\Sigma\widehat{\Omega} - S_1 + S_1 - \Sigma_1)\|_{\max} \\
&= \|\Sigma^{-1}(\Sigma\widehat{\Omega} - S\widehat{\Omega} + S\widehat{\Omega} - S_1) + \Sigma^{-1}(S_1 - \Sigma_1)\|_{\max} \\
&\leq \|(I_d - \Sigma^{-1}S)\widehat{\Omega}\|_{\max} + \|\Sigma^{-1}(S\widehat{\Omega} - S_1)\|_{\max} + \|\Sigma^{-1}(S_1 - \Sigma_1)\|_{\max} \\
&\leq \|\Sigma^{-1}\|_1 \|(\Sigma - S)\widehat{\Omega}\|_{\max} + \|\Sigma^{-1}\|_1 \|S\widehat{\Omega} - S_1\|_{\max} + \|\Sigma^{-1}\|_1 \|S_1 - \Sigma_1\|_{\max} \\
&\leq \|\Sigma^{-1}\|_1 (\|A_1\|_1 \zeta_1 + \lambda_0 + \zeta_2) \\
&= 2\lambda_0 \|\Sigma^{-1}\|_1.
\end{aligned}
$$

Let $\lambda_1$ be a threshold level and we define

$$s_1 = \max_{1 \leq j \leq d} \sum_{i=1}^{d} \min\{|(A_1)_{ij}|/\lambda_1, 1\}, \qquad T_j = \{i : |(A_1)_{ij}| \geq \lambda_1\}.$$

We have, with probability no smaller than $1 - 14d^{-1}$, for all $j \in \{1, \ldots, d\}$,

$$
\begin{aligned}
\|\widehat{\Omega}_{*,j} - (A_1)_{*,j}\|_1 &\leq \|\widehat{\Omega}_{T_j^c,j}\|_1 + \|(A_1)_{T_j^c,j}\|_1 + \|\widehat{\Omega}_{T_j,j} - (A_1)_{T_j,j}\|_1 \\
&= \|\widehat{\Omega}_{*,j}\|_1 - \|\widehat{\Omega}_{T_j,j}\|_1 + \|(A_1)_{T_j^c,j}\|_1 + \|\widehat{\Omega}_{T_j,j} - (A_1)_{T_j,j}\|_1 \\
&\leq \|(A_1)_{*,j}\|_1 - \|\widehat{\Omega}_{T_j,j}\|_1 + \|(A_1)_{T_j^c,j}\|_1 + \|\widehat{\Omega}_{T_j,j} - (A_1)_{T_j,j}\|_1 \\
&\leq 2\|(A_1)_{T_j^c,j}\|_1 + 2\|\widehat{\Omega}_{T_j,j} - (A_1)_{T_j,j}\|_1 \\
&\leq 2\|(A_1)_{T_j^c,j}\|_1 + 4\lambda_0 \|\Sigma^{-1}\|_1 |T_j| \\
&\leq (2\lambda_1 + 4\lambda_0 \|\Sigma^{-1}\|_1) s_1.
\end{aligned}
$$

Suppose $\max_j \sum_{i=1}^d |(A_1)_{ij}|^q \leq s$ and setting $\lambda_1 = 2\lambda_0 \|\Sigma^{-1}\|_1$, we have

$$\lambda_1 s_1 = \max_{1 \leq j \leq d} \sum_{i=1}^{d} \min\{|(A_1)_{ij}|, \lambda_1\} \leq \lambda_1 \max_{1 \leq j \leq d} \sum_{i=1}^{d} \min\{|(A_1)_{ij}|^q/\lambda_1^q, 1\} \leq \lambda_1^{1-q} s.$$

Therefore, we have

$$\|\widehat{\Omega}_{*,j} - (A_1)_{*,j}\|_1 \leq 4\lambda_1 s_1 \leq 4\lambda_1^{1-q} s = 4s(2\lambda_0 \|\Sigma^{-1}\|_1)^{1-q}.$$

Noting that when the lag of the time series $p = 1$, by definition in (3.6), we have $\widehat{\Omega} = \widehat{A}_1$. This completes the proof. $\square$

## A.2 Proof of the Rest Results

*Proof of Corollary 4.4.* Corollary 4.4 directly follows from Theorem 4.1, so its proofs is omitted. $\square$

*Proof of Corollary 4.5.* Using the generating model described in Equation (2.1), we have

$$\|X_{T+1} - \widehat{A}_1^{\mathrm{T}} X_T\|_\infty = \|(A_1^{\mathrm{T}} - \widehat{A}_1^{\mathrm{T}}) X_T + Z_{T+1}\|_\infty$$
$$\leq \|A_1^{\mathrm{T}} - \widehat{A}_1^{\mathrm{T}}\|_\infty \|X_T\|_\infty + \|Z_{T+1}\|_\infty$$
$$= \|A_1 - \widehat{A}_1\|_1 \|X_T\|_\infty + \|Z_{T+1}\|_\infty$$

Using Lemma B.2 in Appendix B, we have

$$\mathbb{P}(\|X_T\|_\infty \leq (\Sigma_{\max} \cdot \alpha \log d)^{1/2}, \|Z_{T+1}\|_\infty \leq (\Psi_{\max} \cdot \alpha \log d)^{1/2}) \geq 1 - 2(d^{\alpha/2-1}\sqrt{\pi/2 \cdot \alpha \log d})^{-1}.$$

This, combined with Theorem 4.1, gives Equation (4.6). $\square$

*Proof of Corollary 4.6.* Similar as the proof in Corollary 4.5, we have

$$\|X_{T+1} - \bar{A}_1^{\mathrm{T}} X_T\|_2 = \|(A_1^{\mathrm{T}} - \bar{A}_1^{\mathrm{T}}) X_T + Z_{T+1}\|_2$$
$$\leq \|A_1 - \bar{A}_1\|_2 \|X_T\|_2 + \|Z_{T+1}\|_2.$$

For any Gaussian random vector $Y \sim N_d(0, Q)$, we have $Y = \sqrt{Q} Y_0$ where $Y_0 \sim N_d(0, I_d)$. Using the concentration inequality for Lipchitz functions of standard Gaussian random vector (see, for example, Theorem 3.4 in Massart (2007)), we have

$$\mathbb{P}(|\|Y\|_2 - E\|Y\|_2| \geq t) = \mathbb{P}(|\|\sqrt{Q}Y_0\|_2 - E\|\sqrt{Q}Y_0\|_2| \geq t)$$
$$\leq 2\exp\left(-\frac{t^2}{2\|Q\|_2}\right). \tag{A.8}$$

Here the inequality exploits the fact that for any vectors $x, y \in \mathbb{R}^d$,

$$|\|\sqrt{Q}x\|_2 - \|\sqrt{Q}y\|_2| \leq \|\sqrt{Q}(x-y)\|_2 \leq \|\sqrt{Q}\|_2 \|x-y\|_2,$$

and accordingly the function $x \to \|\sqrt{Q}x\|_2$ has the Lipschitz norm no greater than $\sqrt{\|Q\|_2}$. Using Equation (A.8), we then have

$$\mathbb{P}(\|X_T\|_2 \leq \sqrt{2\|\Sigma\|_2 \log d} + E\|X_T\|_2, \|Z_{T+1}\|_2 \leq \sqrt{2\|\Psi\|_2 \log d} + E\|Z_{T+1}\|_2) \geq 1 - 4d^{-1}.$$

Finally, we have

$$(E\|Y\|_2)^2 \leq E\|Y\|_2^2 = \mathrm{tr}(Q).$$

Combined with Theorem 4.1 and the fact that $\|A_1 - \bar{A}_1\|_2 \leq \|A_1 - \bar{A}_1\|_1$, we have the desired result. $\square$

*Proof of Theorem 4.7.* Theorem 4.7 follows from the connection between autoregressive model with lag 1 and lag $p$ shown in (2.3). The proof technique is similar to that of Theorem 4.1, thus is omitted. $\square$

# B  Supporting Lemmas

**Lemma B.1** (Negahban and Wainwright (2011))**.** *Suppose that $Y \sim N_T(0, Q)$ is a Gaussian random vector. We have, for $\eta > 2T^{-1/2}$,*

$$\mathbb{P}\Big\{\big|\|Y\|_2^2 - E(\|Y\|_2^2)\big| > 4T\eta\|Q\|_2\Big\} \leq 2\exp\Big\{-T(\eta - 2T^{-1/2})^2/2\Big\} + 2\exp(-T/2).$$

*Proof.* This can be proved by first using the concentration inequality for the Lipchitz functions $\|Y\|_2$ of Gaussian random variables $Y$. Then combining with the result

$$\|Y\|_2^2 - E(\|Y\|_2^2) = (\|Y\|_2 - E\|Y\|_2) \cdot (\|Y\|_2 + E\|Y\|_2),$$

we have the desired concentration inequality. □

**Lemma B.2.** *Suppose that $Z = (Z_1, \ldots, Z_d)^\mathrm{T} \in N_d(0, Q)$ is a Gaussian random vector. Letting $Q_{\max} := \max_i(Q_{ii})$, we have*

$$\mathbb{P}\{\|Z\|_\infty > (Q_{\max} \cdot \alpha \log d)^{1/2}\} \leq \Big(d^{\alpha/2-1}\sqrt{\pi/2 \cdot \alpha \log d}\Big)^{-1}.$$

*Proof.* Simply using the Gaussian tail probability, we have

$$\mathbb{P}(\|Z\|_\infty > t) \leq \sum_{i=1}^d \mathbb{P}(|Z_i| \cdot Q_{ii}^{-1/2} > t \cdot Q_{ii}^{-1/2}) \leq \sum_{i=1}^d \frac{2\exp(-t^2/2Q_{ii})}{t \cdot Q_{ii}^{-1/2} \cdot \sqrt{2\pi}} \leq \frac{2d\exp(-t^2/2Q_{\max})}{t \cdot Q_{\max}^{-1/2} \cdot \sqrt{2\pi}}.$$

Taking $t = (Q_{\max} \cdot \alpha \log d)^{1/2}$ into the upper equation, we have the desired result. □